\documentclass[journal]{IEEEtran}
\usepackage{bibentry}
\usepackage{algorithm}
\usepackage{algorithmic}
\usepackage{color}
\usepackage{multirow}
\usepackage{bbding}
\usepackage{amssymb}
\usepackage{pifont}
\usepackage[table]{xcolor}
\usepackage{colortbl}
\usepackage{booktabs}
\usepackage{footnote}
\usepackage{amsfonts}
\usepackage{bbm}
\usepackage{cite}
\ifCLASSINFOpdf
  \usepackage[pdftex]{graphicx}
  \graphicspath{{./graph/}}
  \DeclareGraphicsExtensions{.pdf,.jpeg,.png}
\else
\fi
\usepackage{amsmath}
\usepackage{algorithmic}
\usepackage{array}
\usepackage{fixltx2e}
\usepackage{stfloats}
\usepackage{url}
\begin{document}
%
% paper title
% Titles are generally capitalized except for words such as a, an, and, as,
% at, but, by, for, in, nor, of, on, or, the, to and up, which are usually
% not capitalized unless they are the first or last word of the title.
% Linebreaks \\ can be used within to get better formatting as desired.
% Do not put math or special symbols in the title.
\title{Industrial Scene Text Detection with Refined Feature-attentive Network}
%
%
% author names and IEEE memberships
% note positions of commas and nonbreaking spaces ( ~ ) LaTeX will not break
% a structure at a ~ so this keeps an author's name from being broken across
% two lines.
% use \thanks{} to gain access to the first footnote area
% a separate \thanks must be used for each paragraph as LaTeX2e's \thanks
% was not built to handle multiple paragraphs
%

% \author{Michael~Shell,~\IEEEmembership{Member,~IEEE,}
%         John~Doe,~\IEEEmembership{Fellow,~OSA,}
%         and~Jane~Doe,~\IEEEmembership{Life~Fellow,~IEEE}% <-this % stops a space
% \thanks{M. Shell was with the Department
% of Electrical and Computer Engineering, Georgia Institute of Technology, Atlanta,
% GA, 30332 USA e-mail: (see http://www.michaelshell.org/contact.html).}% <-this % stops a space
% \thanks{J. Doe and J. Doe are with Anonymous University.}% <-this % stops a space
% \thanks{Manuscript received April 19, 2005; revised August 26, 2015.}}

\author{Tongkun~Guan,~Chaochen~Gu,~Changsheng~Lu,~Jingzheng~Tu,~Qi~Feng,~Kaijie~Wu,~and~Xinping~Guan,~\IEEEmembership{~Fellow,~IEEE}% <-this % stops a space
\thanks{This work is supported by the National Key Research and Development Project of China No. 2019YFB1706602, Chinese Ministry of Education Research Found on
Intelligent Manufacturing NO. MCM20180703. (\emph{Corresponding author: Chaochen Gu}).}
\thanks{T. Guan, C. Gu, J. Tu, Q. Feng, K. Wu, and X. Guan are with the Department of Automation, Shanghai Jiao Tong University, Key Laboratory of System Control and Information Processing, Ministry of Education of China, and Shanghai Engineering Research Center of Intelligent Control and Management, Shanghai 200240, China (e-mail: \{gtk0615, jacygu, tujingzheng, fengqi, kaijiewu, xpguan\}@sjtu.edu.cn).}
\thanks{C. Lu is with the College of Engineering and Computer Science, The Australian National University, Canberra ACT 2600, Australia (e-mail: ChangshengLuu@gmail.com).}% <-this % stops a space
%\thanks{Manuscript received April 19, 2005; revised September 17, 2014.}
\thanks{Copyright © 2022 IEEE. Personal use of this material is permitted. However, permission to use this material for any other purposes must be obtained from the IEEE by sending an email to pubs-permissions@ieee.org.}
}
% note the % following the last \IEEEmembership and also \thanks - 
% these prevent an unwanted space from occurring between the last author name
% and the end of the author line. i.e., if you had this:
% 
% \author{....lastname \thanks{...} \thanks{...} }
%                     ^------------^------------^----Do not want these spaces!
%
% a space would be appended to the last name and could cause every name on that
% line to be shifted left slightly. This is one of those "LaTeX things". For
% instance, "\textbf{A} \textbf{B}" will typeset as "A B" not "AB". To get
% "AB" then you have to do: "\textbf{A}\textbf{B}"
% \thanks is no different in this regard, so shield the last } of each \thanks
% that ends a line with a % and do not let a space in before the next \thanks.
% Spaces after \IEEEmembership other than the last one are OK (and needed) as
% you are supposed to have spaces between the names. For what it is worth,
% this is a minor point as most people would not even notice if the said evil
% space somehow managed to creep in.

% The paper headers
\markboth{Journal of \LaTeX\ Class Files,~Vol.~14, No.~8, August~2021}%
{Shell \MakeLowercase{\textit{et al.}}: Industrial Scene Text Detection with Refined Feature-attentive Network}
% The only time the second header will appear is for the odd numbered pages
% after the title page when using the twoside option.
% 
% *** Note that you probably will NOT want to include the author's ***
% *** name in the headers of peer review papers.                   ***
% You can use \ifCLASSOPTIONpeerreview for conditional compilation here if
% you desire.

% If you want to put a publisher's ID mark on the page you can do it like
% this:
%\IEEEpubid{0000--0000/00\$00.00~\copyright~2015 IEEE}
% Remember, if you use this you must call \IEEEpubidadjcol in the second
% column for its text to clear the IEEEpubid mark.

% use for special paper notices
%\IEEEspecialpapernotice{(Invited Paper)}

% make the title area
\maketitle

% As a general rule, do not put math, special symbols or citations
% in the abstract or keywords.
\begin{abstract}
  Detecting the marking characters of industrial metal parts remains challenging due to low visual contrast, uneven illumination, 
  corroded surfaces, and cluttered background of metal part images. 
  Affected by these factors, bounding boxes generated by most existing methods could not locate low-contrast text areas very well. 
  %locate low-contrast text areas inaccurately.
  In this paper, we propose a refined feature-attentive network (RFN) to solve the inaccurate localization problem. 
  Specifically, we first design a parallel feature integration mechanism to construct an adaptive feature representation 
  from multi-resolution features, which enhances the perception of multi-scale texts at each scale-specific level to generate a 
  high-quality attention map. 
  Then, an attentive proposal refinement module is developed by the attention map to rectify the location deviation of candidate boxes. 
  Besides, a re-scoring mechanism is designed to select text boxes with the best rectified location. 
  To promote the research towards industrial scene text detection, we contribute two industrial scene text datasets, including a total of 102156 images and 1948809 text instances with 
  various character structures and metal parts. 
  Extensive experiments on our dataset and four public datasets demonstrate that our proposed method achieves the state-of-the-art performance. 
  Both code and dataset are available at: https://github.com/TongkunGuan/RFN.
\end{abstract}

% Note that keywords are not normally used for peerreview papers.

% For peer review papers, you can put extra information on the cover
% page as needed:
% \ifCLASSOPTIONpeerreview
% \begin{center} \bfseries EDICS Category: 3-BBND \end{center}
% \fi
%
% For peerreview papers, this IEEEtran command inserts a page break and
% creates the second title. It will be ignored for other modes.
\IEEEpeerreviewmaketitle

\section{Introduction}
\begin{figure}[t]
  \centering
  \includegraphics[width=3.5in]{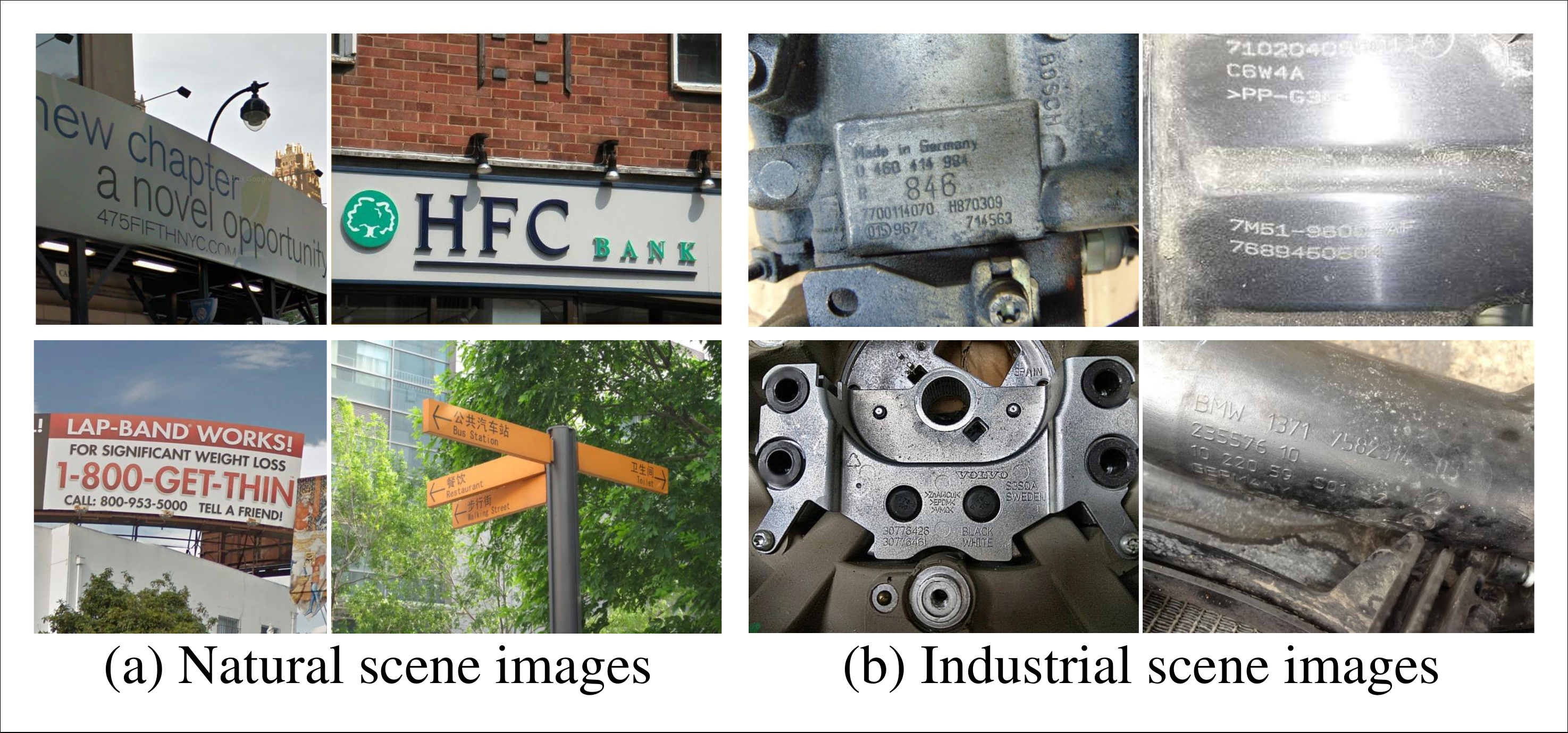}
  \caption{Visual comparisons between different scene text detection datasets. (a) Natural scene images, (b) Industrial scene images.}
  \label{Figs.NSTvsIST}
\end{figure}

% The very first letter is a 2 line initial drop letter followed
% by the rest of the first word in caps.
% 
% form to use if the first word consists of a single letter:
% \IEEEPARstart{A}{demo} file is ....
% 
% form to use if you need the single drop letter followed by
% normal text (unknown if ever used by the IEEE):
% \IEEEPARstart{A}{}demo file is ....
% 
% Some journals put the first two words in caps:
% \IEEEPARstart{T}{his demo} file is ....
% 
% Here we have the typical use of a "T" for an initial drop letter
% and "HIS" in caps to complete the first word.
\IEEEPARstart{T}{he} goal of text detection is to localize the text regions with bounding boxes, 
which mainly includes horizontal texts, multi-oriented texts, and curved texts in various scenarios. 
With the advent of laser marking technology, many metal parts are marked with Latin characters and Arabic numerals to record the serial number,  production date, and other product information. 
Detecting these texts plays an increasingly important role in intelligent industrial manufacturing, which is conducive to improving the 
assembly speed of industrial production lines and the efficiency of logistics transmission in the industrial scene. 
Compared with the natural scene text detection (\emph{e.g.}, traffic signs, shopping mall trademarks, and billboards), 
industrial scene text detection has low visual contrast, corroded surfaces, and complex backgrounds. 
Thus, these characteristics pose greater challenges to industrial text detection.
Specifically, the differences between industrial scenes and natural scenes are shown in Fig.\,\,\ref{Figs.NSTvsIST}.

\begin{figure}[t]
  \centering
  \includegraphics[width=3.5in]{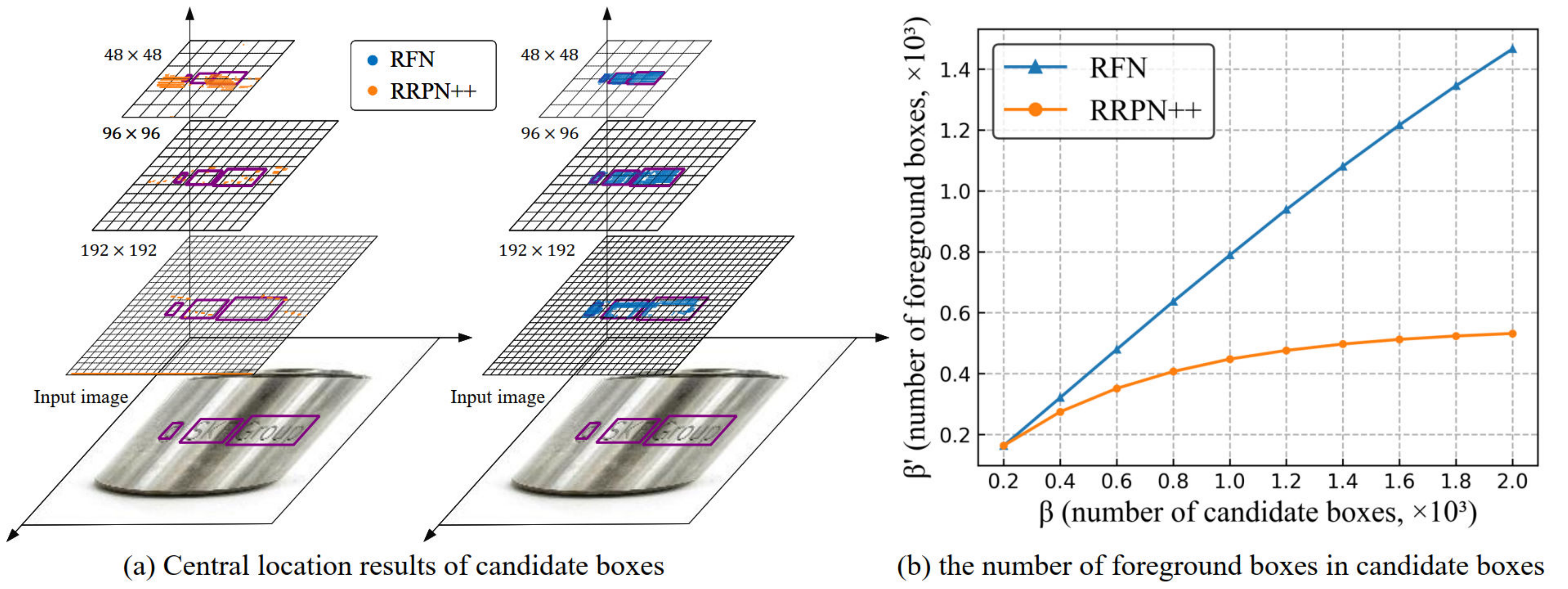}
  \caption{
  Central location results of candidate boxes generated by two-stage detectors (\emph{i.e.}, RRPN++, RFN(ours)) on an industrial image in (a).
  We also calculate the average number of foreground boxes on our MPSC dataset under the same number of candidate boxes in (b).
  Specifically, our method contains more foreground boxes at the same number of candidate boxes compared to RRPN++.
  Note that a candidate box is defined as foreground box if its center point locates in a groundtruth.
  }
  \label{Figs.RPNvsRFN}
\end{figure}

Conventional scene text detection (STD) methods firstly extract regions of interest using shape detectors \cite{Arc:2019Arc, Circle:2018Circle} and then search the text boxes, while the existing ones are mainly based on deep neural networks and consist of three categories:  
segmentation-based methods, regression-based methods, and a combination of segmentation and regression methods. 
Most segmentation-based text detection methods\cite{FCN:Long2015Fully,CCTN:he2016accurate,Pixellink:deng2018pixellink, 
Border:8237797, Shape-Aware_Embedding,PSENET:wang2019shape,DB:2020Real,TCSVT:9214994} adopt semantic segmentation to perform pixel-level classification (\emph{i.e.}, text/non-text prediction) and group these pixels belonging to text to generate bounding boxes. 
However, the text edges of metal parts are less clear than natural scene texts. 
A segmentation network that eliminates background noise causes misclassification of foreground in low-contrast industrial images, which leads to inaccurate localization during post-processing.
Regression-based text detection methods\cite{Textboxes:liao2016textboxes,Textboxes++:Liao2018TextBoxes,DMPNet:liu2017deep, RRPN:2018Arbitrary, 
FEN:zhang2018feature, LOMO:zhang2019look, R3Det:yang2019r3det, EAST:zhou2017east, BDN:liu2020exploring, TCSVT:8869778} mainly establish geometry metrics on text boxes and calculate regression loss for localizing texts. 
These methods use one-stage or two-stage detectors to implement text localization. 
The methods based on a one-stage detector run faster but have lower accuracy. 
The methods based on a two-stage detector generate preliminary detection boxes by a region proposal network (RPN)\cite{fasterRCNN:2017Faster} 
and then select better boxes to feed into a refinement network according to confidence scores. 
Although the two-stage detectors correct the location of each box, the candidate box quality on metal parts still needs to be improved.
As shown in Fig.\,\,\ref{Figs.RPNvsRFN}, we visualize the center locations of these candidate boxes of RRPN++\cite{RRPN++:ma2020rrpn++} and RFN (ours).
The center points of candidate boxes generated by RRPN++ deviate from the text groundtruth in industrial scenes, 
which increases the difficulty of location correction. 

Therefore we propose a refined feature-attentive network (RFN) by utilizing more foreground information 
to improve localization accuracy.
Specifically, we first design a segmentation-based foreground-focus module (SFF) to generate a high-quality attention map with more 
detailed information of insensitive character areas. The SFF module adaptively integrates multi-resolution features to enhance the perception of multi-scale text features at each scale-specific layer. 
Then, an attentive proposal refinement module (APR) applies the attention map to construct high-quality foreground boxes, which generates discriminative classification and regression results for localization correction.
Finally, experiments demonstrate that our method achieves the state-of-the-art performance on the MPSC dataset and robustly detects horizontal texts, multi-oriented texts, and multi-language texts on the MSRA-TD500\cite{MSRA:article}, USTB-SV1K\cite{USTB-SV1K}, ICDAR2013\cite{ICDAR2013:2013ICDAR}, and ICDAR2017-MLT\cite{ICDAR2017-MLT:2017ICDAR2017} public datasets. 

In addition, we contribute a benchmark dataset on metal parts for industrial text detection.
To the best of our knowledge, it is the first industrial text detection benchmark dataset.
Specifically, we build a metal part surface character dataset (MPSC) for industrial scenes and synthesize a SynthMPSC dataset on metal images to expand the types and qualities. 
The proposed MPSC dataset includes common challenges in natural scenes, \emph{e.g.}, multiple orientations,
multiple scales, and complex background, and poses great challenges in industrial scenes, \emph{e.g.}, corroded surfaces, low visual contrast, and uneven illumination.
In summary, the main contributions of this paper are three-fold:
\begin{itemize}
  \item[1)] 
  We propose a refined feature-attentive network (RFN) for industrial text detection, which focuses on foreground information and generates high-quality text boxes to improve the localization accuracy of metal parts.
  \item [2)]
  In our RFN, a segmentation-based foreground-focus module (SFF) and an attentive proposal refinement module (APR) are proposed. 
  The SFF module guides the framework to focus on the text features of metal parts by learning adaptive feature representations.
  The APR module is developed to construct high-quality foreground boxes for text localization.
  \item [3)] 
  We establish a challenging large-scale industrial text detection benchmark dataset (MPSC) and synthesize a SynthMPSC dataset based on real-world metal images. The MPSC dataset is the first industrial text detection benchmark dataset.
\end{itemize}

\begin{figure*}[t]
  \centering
  \includegraphics[width=6.8in]{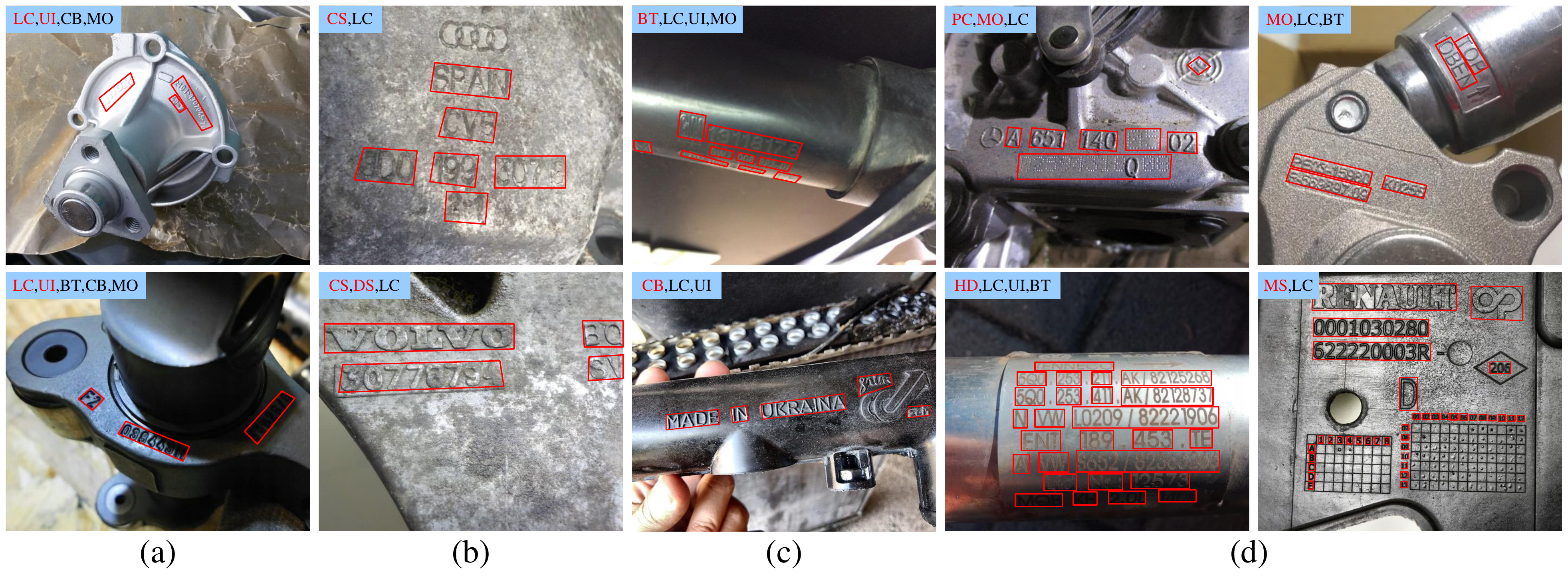}
  \caption{Examples of MPSC dataset. 
  Our datasets bring many new challenges affected by the following factors: 
(a) \emph{Metal properties}, industrial text detectors should consider the challenges of low visual contrast (LC) and uneven illumination (UI) on parts due to metal materials.
(b) \emph{Industrial circumstances}, the parts appear corroded surfaces (CS) and dirty surfaces (DS) due to the influence of weather and humidity in the production workshop.
(c) \emph{Scene noise}, unconstrained motion shooting produces blur texts (BT) and introduces an industrial complex background (CB).
(d) \emph{Artificial design}, product information is mainly presented on metal parts in multiple forms, including multi-direction (MO), multi-scale (MS), high-density (HD), and polymorphic characters (PC). 
  The blue region in the upper left of each image depicts the corresponding challenges. 
  Each image provides accurate text transcription and clockwise ground-truth boxes. }
  \label{Figs.MPSC}
\end{figure*}
\section{Related Work}
In this section, we review the development of text detection methods.
Early methods adopt Fourier and Laplacian transform \cite{TCSVT:5585735}, SVM\cite{TCSVT:8320866}, 
connected components analysis (CCA)\cite{2014Text,MSRA:article,2013Robust}, 
maximally stable extremal regions (MSER)\cite{TCSVT:7944571} and sliding window (SW) 
based classification \cite{2011AdaBoost,2012End,2011Text} methods to implement text localization tasks.
However, the above methods process text components in a bottom-up order, with long and slow pipelines. 
Later, inspired by general object detection, they utilize deep convolutional neural networks (CNNs) to generate a variety of text geometric metrics, 
such as bounding boxes, pixel-level masks, contour points, text centerlines, etc. 
These methods can be roughly divided into three categories: 
regression-based methods, segmentation-based methods, and combination segmentation and regression methods.
\subsection{Regression-based Methods}
Regression-based text detection methods \cite{Textboxes:liao2016textboxes,Textboxes++:Liao2018TextBoxes,DMPNet:liu2017deep, RRPN:2018Arbitrary, 
FEN:zhang2018feature, LOMO:zhang2019look, R3Det:yang2019r3det, EAST:zhou2017east, BDN:liu2020exploring, TCSVT:8869778} predict the offsets from key elements and decode them into bounding boxes. 
Inspired by SSD\cite{SSD:liu2016ssd}, methods utilizing the pre-defined anchors (key element) simplify the detection pipeline, 
which are end-to-end trainable. 
By adding six text-box layers based on SSD, Liao \emph{et al.} propose Textboxes++\cite{Textboxes++:Liao2018TextBoxes} to predict the offsets 
from the pre-defined anchors composed of different aspect-ratios and scales. Similarly, Wang \emph{et al.}\cite{DMPNet:liu2017deep} 
first design prior quadrilateral sliding windows for locating multi-oriented texts, which are different from horizontal sliding windows. 
Ma \emph{et al.}\cite{RRPN:2018Arbitrary} further add an angle specification into anchor strategy to generate rotating region proposals, which matches 
text instances of arbitrary orientations. 
However, single-stage detectors generate increasing failure examples on the cluttered background, 
which degrades text detection performance.
Consequently, different refinement methodologies are adopted to optimize localization results. 
Similar to two-stage object detection methods, text detectors
\cite{RRPN:2018Arbitrary,FEN:zhang2018feature} extract text features from the proposals generated by an RPN-like mechanism and then adopt 
ROI pooling\cite{fasterRCNN:2017Faster} or RoIAlign\cite{MaskR-CNN:2017Mask} to obtain fixed-scale feature maps. The branches of boxes 
classification and boxes regression are finally utilized to correct the localization results of each proposal. On the basis of two-stage 
detectors, Zhang \emph{et al.}\cite{LOMO:zhang2019look} and Yang \emph{et al.}\cite{R3Det:yang2019r3det} propose iterative refinement modules 
to implement text detection correction and improve localization precision. 
Moreover, Zhou \emph{et al.}\cite{EAST:zhou2017east} adopt an anchor-free strategy to realize geometry-aware localization.
It generates boxes by predicting box edge distances from the current pixel (key element) 
to the minimal bounding rectangle of its text instances, and then combine the score map to detect the arbitrary-oriented text. 
\subsection{Segmentation-based Methods}
Inspired by FCN\cite{FCN:Long2015Fully}, many segmentation-based methods \cite{FCN:Long2015Fully,CCTN:he2016accurate,Pixellink:deng2018pixellink, 
Border:8237797, Shape-Aware_Embedding,PSENET:wang2019shape,DB:2020Real,TCSVT:9214994} adopt semantic segmentation and instance segmentation for text 
detection. He \emph{et al.}\cite{CCTN:he2016accurate} adopt cascaded convolutional networks to implement coarse-to-fine 
segmentation based on text instances and the centerline of text lines for text localization. Deng \emph{et al.}\cite{Pixellink:deng2018pixellink} 
segment text/non-text by linking pixels in the same instance and conduct post-processing to extract text bounding boxes without location 
regression. Wu \emph{et al.} \cite{Border:8237797} implement text detection by introducing a border class, 
and a lightweight FCN is applied to cast each pixel into three categories: text, border, and background. 
Tian \emph{et al.}\cite{Shape-Aware_Embedding} optimize a shape-aware loss to distinguish the pixels among different text instances by embedding a space vector for each pixel.
Specifically, it maximizes the Euclidean distances of pixel embedding vectors 
from different text instances and minimizes those belonging to the same instance. Wang \emph{et al.}\cite{PSENET:wang2019shape} gradually 
expand the minimal scale kernel size and increase the segmentation area for detecting text instances of arbitrary shapes. 
Liao \emph{et al.}\cite{DB:2020Real} develop a differentiable binarization (DB) algorithm for the segmentation network, which performs binarization with an approximate step 
function and makes the process end-to-end trainable. Instead of setting the fixed thresholds, the segmentation network adds an adaptive 
threshold map per image to provide a highly robust text feature map.
\subsection{Combination of Segmentation and Regression Methods}
The combination segmentation and regression methods are proposed to improve the effect of text detection lately. 
He \emph{et al.}\cite{SSTD:he2017single} exploit a regional attention mechanism to predict locations and scores of text boxes. 
Based on Mask R-CNN \cite{MaskR-CNN:2017Mask}, Xie \emph{et al.}\cite{SPCNET:xie2018scene} merge a text-context module into joint multi-scale 
pyramid features in order to suppress false alarms and reduce the number of false-positive boxes. 
Similarly, Huang \emph{et al.}\cite{Mask_PAN:huang2019mask} update feature extraction network derived from Pyramid Attention Network
\cite {PAN:li2018pyramid}, and add a text mask prediction branch that detects curved texts.
In addition, Yang \emph{et al.}\cite{Inception:yang2018inceptext} and Dai \emph{et al.}\cite{FTSN:2018Fused} use 
a fully convolutional instance-aware semantic segmentation (FCIS)\cite{FCIS:li2017fully} 
method to guide the prediction of three text-related elements: mask, class, and box, by generating an instance-aware segmentation 
map. Wang \emph{et al.}\cite{ContourNet:2020ContourNet} propose an Adaptive-RPN with a scale-insensitive metric 
to accurately generate proposal bounding boxes, and then add contour characteristic of text regions by executing the convolution operation 
in two orthogonal directions to locate texts with arbitrary shapes.

\section{MPSC \& SynthMPSC Dataset}
The publicly available text detection datasets are mainly taken from natural scenes, and no industrial datasets can be explored and 
researched in the community. 
In this section, we establish a benchmark dataset (Metal Part Surface Character Dataset, MPSC) to promote in-depth research on text detection in industrial scenes. 
Specifically, our dataset spans many challenges affected by four factors (\emph{i.e.}, metal
properties, industrial circumstances, scene noise, artificial design) as shown in Fig.\,\,\ref{Figs.MPSC}. For instance, low visual contrast, uneven illumination, 
corroded surfaces, dirty part surfaces, blurred texts, clutter background, polymorphic characters, and multi-orientations.
Moreover, we build an artificial metal part text dataset (Synthesized Metal Part Surface Character Dataset, SynthMPSC) 
by synthesizing characters with real-world metal part images.
\subsection{MPSC Dataset}
By fully considering different types and styles of characters and metal parts, we collect a metal part surface character (MPSC) 
dataset.
Specifically, 3194 images are constructed into the MPSC dataset, including 2555 training images and 639 testing images.  
\subsubsection{Dataset Construction}
We perform industrial data deduplication, data cleaning, and data labelling on the collected images for three months, 
to promote the industrial application of text detection to a new stage.
First, each image needs to be scrutinized that simple images and unqualified images are removed.
Refer to ICDAR 2015 incidental text dataset\cite{ICDAR2015:2015ICDAR}, qualified images are then labelled with quadrilaterals at word-level 
where the four corners must be arranged clockwise. Finally, three rounds of inspections are implemented to reduce manual labelling errors.
\subsubsection{Dataset Property Analysis}
The MPSC dataset provides high-quality ground-truth boxes and text transcriptions. Most of them contain special combination rules that are 
different from the legal spelling of words, such as "AlSi9Cu3", "D151C-050506", and "7M121". 
Each label has a specific implication, which embodies the signification of character encoding in industrial scenes. 
In addition, sufficient statistical results are calculated to show more information about the MPSC dataset.
First, the number of characters per text instance is distributed between 1 to 31, with the majority ranging from 2 to 7, and 5.5 is the average. 
Second, the aspect ratios greater than 1 account for 70.6\% of all text instances, while 1 and other aspect ratios account for 19.2\% and 
10.2\%, respectively. 
Finally, the width of 92.9\% of text instances is no more than 40\% of the image width, while the height of 97.8\% of text instances is less than 
20\% of the image height. 
Therefore, the area of most text instances is less than 8\% of the image area. 
\begin{figure}[t]
  \centering
  \includegraphics[width=3.5in]{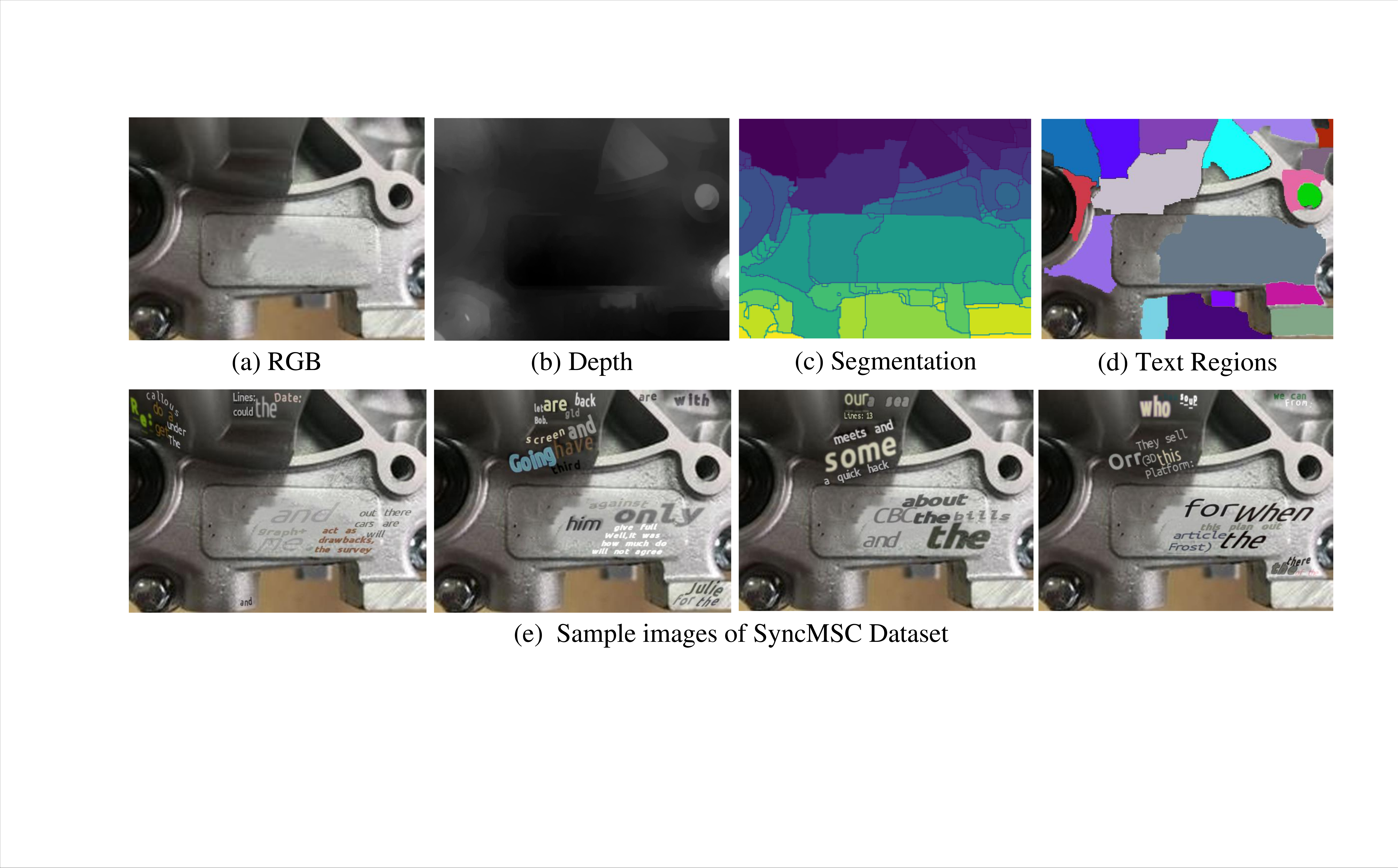}
  \caption{The generation procedure of SynthMPSC dataset. 
  (Top) Four flowcharts generated by the algorithm. 
  The RGB image is first predicted to generate a depth map and a segmentation map, 
  and then the segments suitable for placing texts in the segmentation map are defined as text regions to synthesize characters.
  (Bottom) Some synthetic images on the SynthMPSC dataset.}
  \label{Figs.SyncMPSC}
\end{figure}
\subsection{SynthMPSC Dataset}
\subsubsection{Motivation}
The self-built dataset fully considers the possibility of character structures and metal parts, 
whereas some attributes, \emph{e.g.}, character types, aspect ratios, area ratios, and directions, may exist uneven distribution as other 
public datasets, which limits the capability of sophisticated methods in real-world scenarios. 
Though current transfer neural networks \cite{lu2020deep, lu2018viewpoint, wu2021domain} are promissing to reduce domain shift to real texts, we use a simple synthtext algorithm\cite{Synthtext:2016Synthetic} to generate large batches of synthetic images 
with rich text attributes. 
\subsubsection{Data Synthesis}
We collect 1,153 metal images without characters to synthesize the SynthMPSC dataset containing 98,962 images and 1933234 text instances. 
Fig.\,\,\ref{Figs.SyncMPSC} describes the visualization process of the SynthMPSC dataset.
Specifically, the generation process of the SynthMPSC dataset starts with sampling images and texts. 
First, we select an image without characters that accord with the metal background characteristics, and predict its dense depth map. 
Then, we segment the image into multiple regions by colour and texture cues\cite{cues:Pablo2011Contour}, 
and the areas suitable for text placement are marked with random colouring. 
Finally, we extract the corpus from the Newsgroup20 dataset\cite{Newsgroup:Lang95} and pick the appropriate fonts and colours to 
synthesize images of metal parts with texts. 
\subsection{Comparison with Other Public Datasets}
Public datasets ICDAR2013\cite{ICDAR2013:2013ICDAR}, ICDAR2015\cite{ICDAR2015:2015ICDAR}, MSRA-TD500\cite{MSRA:article}, 
and USTB-SV1K\cite{USTB-SV1K} are widely used in text localization tasks. 
They are taken from natural scenes, including traffic signs, shopping mall trademarks, billboards, etc. 
These texts have relatively clear texts with variable styles and colours against a chaotic background.
For example, merchants want their trademarks to be more colorful and distinctive for attracting customers' attention. 
Advertisers use clear and bright texts to let readers understand product value straightforwardly. 
\subsubsection{ICDAR 2013}
It is widely used to implement text detection and recognition tasks.
With its text instances almost horizontal, images are annotated with rectangular boxes.
\subsubsection{ICDAR 2015} 
It is taken from street-viewed scenes.
The text regions are annotated by 4 vertices of the quadrangle.
\subsubsection{MSRA-TD500}
It uses line-level bounding boxes instead of word-level bounding boxes to annotate labels so that the entire dataset contains many 
large-scale and long text instances. 
These images contain both Chinese texts and English texts. 
\subsubsection{USTB-SV1K}
It contains many low-resolution and blur images.
These images are artificially blurred to a certain extent, and text instances are characterized by multiple orientations, views, and 
perspective distortion.
\begin{table}[t]
  \caption{Statistic Comparison Between Our MPSC and Other Benchmarks}
  \centering
  \scalebox{0.85}{
  \begin{tabular}{ccccccccccccc}
  \toprule
  \multirow{2}{*}{Dataset} & \multicolumn{3}{c}{Image} & \multicolumn{3}{c}{Label} & \multirow{2}{*}{Direction} \\ 
  \cmidrule(r){2-4} \cmidrule(r){5-7} 
                 & train & test & all   & character  & word     & line  \\                                                
  \midrule
  ICDAR 2013     & 229  & 233  & 462    & \checkmark & \checkmark & -                    & Horizontal     \\
  MSRA-TD500     & 300  & 200  & 500    & -          & -          & \checkmark          & Multiple               \\
  ICDAR 2015     & 1000 & 500  & 1500   & -          & \checkmark & -                    & Multiple               \\
  USTB-SV1K      & 500  & 500  & 1000   & -          & -          & \checkmark          & Multiple               \\
  \toprule
  MPSC (ours)          & \textbf{2555} & \textbf{639}  & \textbf{3194}  & -          & \checkmark & -           & Multiple    \\
  SynthMPSC (ours)     & \textbf{98962}& -    & \textbf{98962} &  \checkmark & \checkmark & -                  & Multiple    \\
  \bottomrule
  \end{tabular} 
  }
  \label{tb:001}
\end{table}

The quantitative and statistic comparison results between the proposed dataset and other benchmarks are summarized in Table \ref{tb:001}.
The number of images in our MPSC is 6.91 times \emph{(i.e., 3, 194 vs. 462)}, 6.388 times \emph{(i.e., 3, 194 vs. 500)}, 2.129 times \emph{(i.e., 3, 194 vs. 1500)} and 
3.194 times \emph{(i.e., 3, 194 vs. 1000)} that in ICDAR 2013, MSRA-TD500, ICDAR 2015 and USTB-SV1K, respectively.
Our MPSC dataset is the first industrial text detection benchmark dataset. It covers many challenges from the natural scene \emph{e.g.}, 
multiple orientations, multiple scales, and complex background, and poses great challenges \emph{e.g.} corroded surfaces, 
low contrast, and uneven illumination, in the industrial scene to the state-of-the-art methods.

\section{Refined Feature-attentive Network}
In this section, we propose an industrial text detection method to locate text robustly and effectively. 
First, we summarize the overall structure of our proposed method, then illustrate the details of the SFF, APR, and Re-scoring modules, 
and finally, introduce the loss function of our method. 
\subsection{Overall Pipeline}
\begin{figure*}[htbp]
  \centering
  \includegraphics[width=6.8in]{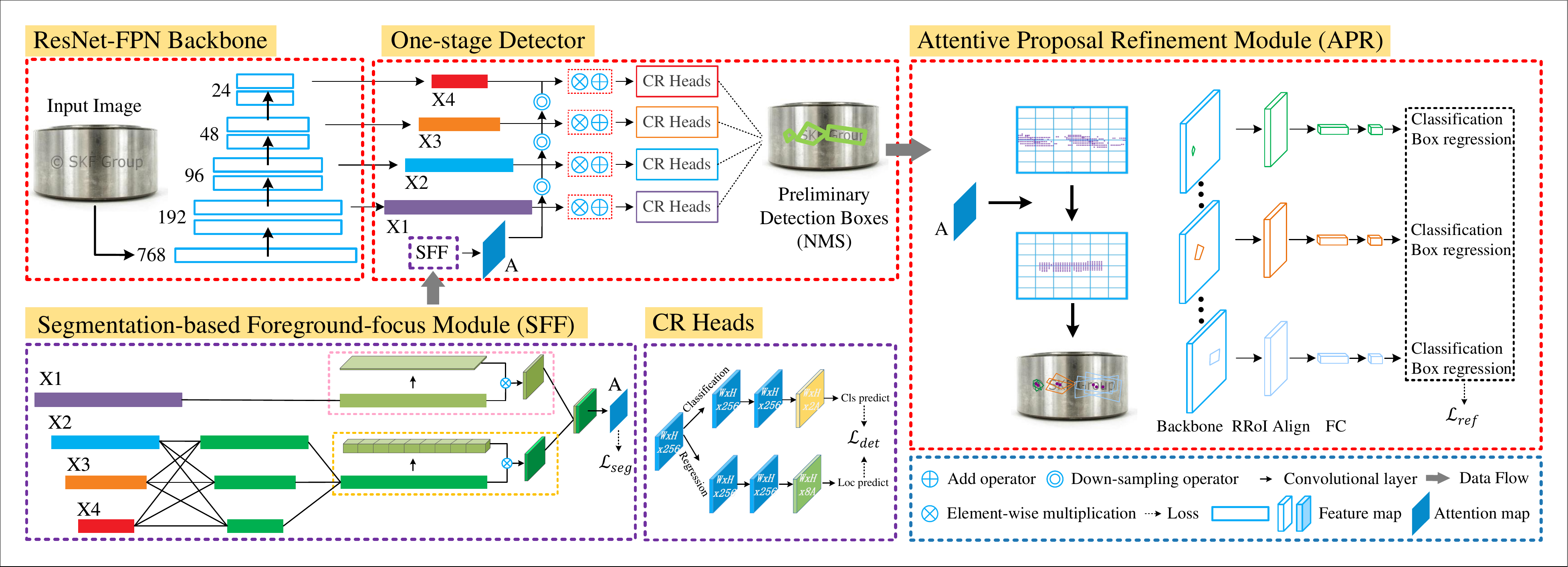}
  \caption{The overall framework of our proposed method. 
  The entire industrial text detection process consists of ``ResNet-FPN Backbone'', ``One-stage Detector'', and ``Attentive Proposal Refinement Module'', 
  shown in the three big red dotted boxes. 
  Firstly,  multi-scale features are extracted from the ResNet-FPN backbone and fused to form an attention map by the segmentation-based 
  foreground-focus module. 
  Then, multi-scale features weighted by the attention map are fed into the classification and regression subnets (``CR Heads'') to predict 
  the preliminary detection boxes.
  After that, the attentive proposal refinement module mines high-quality candidate boxes attached to the foreground to correct location.
  }
  \label{Figs.Network}
\end{figure*}

Our network mainly includes four parts: 
a ResNet-FPN backbone for extracting multi-scale features, a detection branch of classification and regression tasks, 
a semantic segmentation branch for highlighting foreground features, and an attentive proposal refinement module. 
In RFN, we first employ ResNet\cite{ResNet:2016Deep} with 50 layers as a backbone of FPN\cite{FPN:2016Feature}, 
which is used to extract the multi-scale text features. 
Then, we introduce a segmentation-based foreground-focus module to highlight and retain more text area features. 
They are fed into two structure-sharing and parameter-separating sub-networks for obtaining preliminary detection boxes. 
Next, an attentive proposal refinement module corrects the location deviation of candidate boxes in which high-quality preliminary detection boxes are 
selected by combining multi-scale attention maps. 
Finally, we establish a re-scoring mechanism to assess the quality of the correction boxes by combining instance and classification scores. 
We illustrate the specific details of the text detection network RFN in Fig.\,\,\ref{Figs.Network}.
\subsection{Segmentation-based Foreground-focus Module}
The surface of metal parts has a complicated visual context, with similar texture, uneven illumination, and varying character structures. 
Thus a feature extraction network should provide robust feature representations with multi-scale text features on complex metal surfaces. 
However, layers with different resolutions have perception differences in multi-scale text features. 
While a high-scale feature map represents more details to capture small objects, 
the low-scale feature map with more decisive semantic information is usually more suitable for large objects. 
Existing methods mainly adopt the bottom-up integration approach to learn text features, 
which weaken the perception of multi-scale text features at each scale-specific layer and 
may not provide robust feature representations on complex metal surfaces. 
To exchange the information across multi-scale representations, 
we design the following network to enhance the ability to fusion and complementarity between feature layers with different scales.
\subsubsection{Network Design}
We design a novel feature extraction module from complex metal backgrounds, 
adaptively fusing multi-resolution features to enhance the perception of multi-scale text features at each scale-specific layer.
First, we employ the ResNet-FPN backbone\cite{FPN:2016Feature} to extract multi-scale features, which are defined as $\{X_{1},X_{2},X_{3},X_{4}\}$.
With different resolutions $s_i=(h_i,w_i), i=\{1,2,3,4\}$, they are divided into low- (\emph{i.e., $X_1$}) and high- 
(\emph{i.e., $X_2$, $X_3$}, and \emph{$X_4$}) levels to enhance text feature representations in different ways.
It compensates for the deficiencies of the scale-specific layer and enhances the adaptability to complex variations.

For low-level input, we highlight text information to enhance the semantic features. Specifically, the input 
$ X_1 \in R^{h_1\times w_1\times c}$ is first fed into several convolutional layers with BatchNorm and ReLU activation function. 
Followed by average-pooling operation along the channel axis, the foreground response value naturally gets a high accumulation. 
Instead of the sigmoid function, 
the low-level attention map is activated by the exponential operation to expand the difference between the response 
weights of foreground and background due to similar texture. Finally, the attention map is broadcasted and 
element-wise multiplicated with the input images $X_1$ to get low-level response maps $L$.

For high-level input, we design a parallel structure to fuse multi-resolution feature maps by exchanging information mutually. 
Each subnet of the high-level input adaptively learns the features of adjacent subnets, which enriches the spatial 
features and retains more multi-scale text features. Taking $X_i \in R^{h_i\times w_i\times c}, i=\{2,3,4\}$, 
the parallel structure is summarized based on:
\begin{gather}
  Y_k = \sum_{i = 2}^{4}\mathcal{F}(X_i,s_k),
\end{gather}
where $\mathcal{F}(X_i,s_k)$ means $X_i$ is upsampled or down-sampled operation from resolution $s_i$ to $s_k$.
Specifically, upsampling operation refers to $1\times1$ convolutional layers followed by the bilinear sampler, while 
$3\times3$ convolutional layers with the stride of $2$ are implemented for down-sampling.
If $s_i = s_k$, $\mathcal{F}(\cdot)$ represents $3\times3$ convolutional layers without sampling layer.
We fuse multi-resolution $Y_k$ to generate the final high-level response map $H$ in the following:
\begin{gather}
  \begin{aligned}
  H
  & = \varphi(\mathcal{T}(Y_2,s_1) \parallel \mathcal{T}(Y_3,s_1) \parallel \mathcal{T}(Y_4,s_1)),
  \end{aligned}
\end{gather}
where $\mathcal{T}(\cdot )$ refers to upsampling $Y_k$ from resolution $s_k$ to $s_1$, ``$\parallel$'' represents the concatenation 
along the channel axis, and $\varphi(\cdot)$ means adopting the channel-wise attention mechanism assigns large responses 
for foreground features.

\begin{figure}[t]
  \centering
  \includegraphics[width=3.5in]{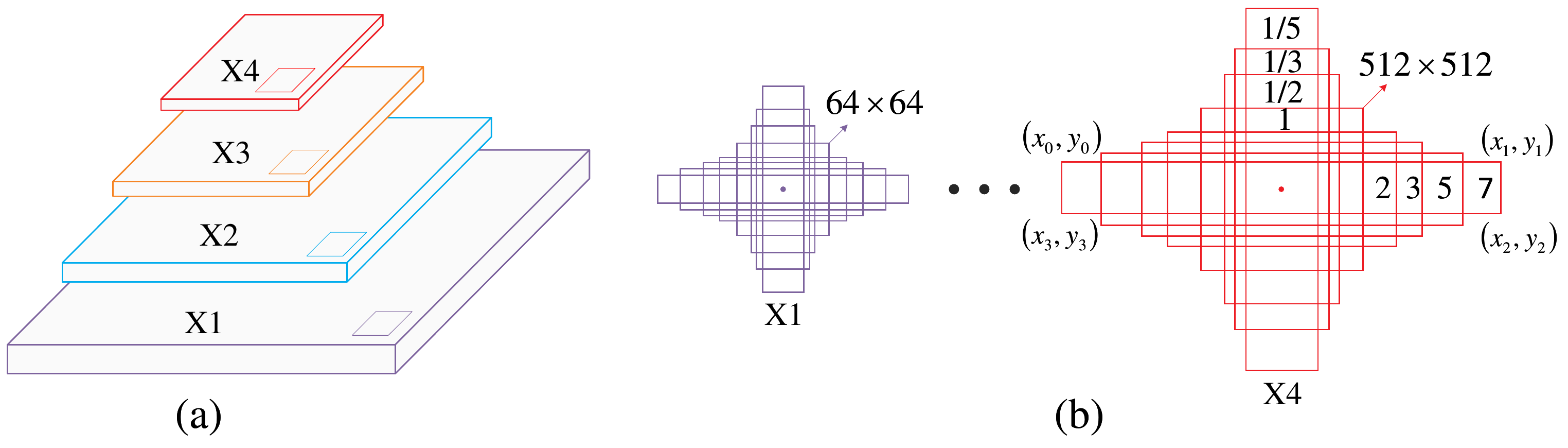}
  \caption{The dafault boxes strategy used in our network. (a) Pyramid network composed of $X_1,X_2,X_3,X_4$. (b) The default boxes with different scales and aspect ratios.}
  \label{Figs.Anchor}
\end{figure}
Multi-level text features (\emph{i.e. Low-level response map $L$, High-level response map $H$}) are then fused to generate an attention map $A$, which provides 
rich and discriminative semantic information and endows higher foreground response values. It guides sub-networks of all levels
to focus on text features. We implement the specific details as follows:
\begin{gather}
  \hat{X_i} =X_i \odot (1+e^{\mathcal{H}(A,s_i)}),
\end{gather}
where $\mathcal{H}(\cdot )$ refers to down-sampling the attention map $A$ from resolution $s_1$ to $s_i$.

Subsequently, each subnet $\hat{X_i}$ is fed into regression and classification branches respectively. 
They adopt the common structure with separate parameters, consisting of four 3x3 convolutional layers, 
followed by a 3x5 convolutional layer for oriented texts.
Based on pre-defined anchors from the generation strategy of Fig.\,\,\ref{Figs.Anchor}, each subnet $\hat{X_i}$ generates a total of 
$h_i \times w_i \times 8$ preliminary detection boxes denoted as $\mathcal{B}_i $, and $h_i \times w_i \times 8$ confidence scores denoted as $\mathcal{S}_i $.
\subsubsection{Segmentation Loss}
We propose a novel loss function to boost the segmentation result of the supervised attention map. 
Unlike the natural scene text, the text edge of metal parts is unclear, and the visual contrast is low, 
which brings challenges to the accurate distinction between foreground and background areas.
Thereby we pay more attention to the foreground and take two original intentions for establishing the attention map's loss mechanism 
in the order of priority.
 a) Include as many text features as possible. b) Minimize the amount of false detection. 
Specifically, assuming that the groundtruth foreground mask is $S_{gt}$, which can be constructed from the quadrilateral bounding boxes. 
We first use the Dice \cite{Dice:2016V} as the auxiliary loss function to deal with the extremely unbalanced positive and negative samples, 
since the text areas of interest occupies only a very small region of the image.
\begin{gather}
  \mathcal{L}_{d} = 1-\frac{2*\sum_{i=1}^{N}(\omega_{i}\omega_{i}^{*})}{\sum_{i=1}^{N}(\omega_{i}) + \sum_{i=1}^{N}(\omega_{i}^{*})},
\end{gather}
where $N$ is the number of pixels in the attention map $A$, 
$\omega_{i}$ and $\omega_{i}^{*}$ are the confidence score of pixel $i$ in $S_{gt}$ and $A$, respectively.
Then the coefficient of false negative and false positive can be calculated as $\mathcal{D}_{a}$ and $\mathcal{D}_{b}$, respectively.
\begin{gather}
  \omega_{d} = \omega_{i} - \omega_{i}^{*},\\
  \mathcal{D}_{a} = \frac{\sum_{i=1}^{N} \mathbbm{1}_{[\omega_{d} \geqslant \frac{1}{2}]} (1-\omega_{i}^{*})}{\sum_{i=1}^{N}(\omega_{i}^{*})},\\
  \mathcal{D}_{b} = \frac{\sum_{i=1}^{N} \mathbbm{1}_{[-\omega_{d} \geqslant \frac{1}{2}]}\omega_{i}^{*}}{\sum_{i=1}^{N}(\omega_{i}^{*})}
\end{gather}
Finally the loss function are designed as follows: 
\begin{gather}
  \mathcal{L}_{g}=
  \begin{cases}
  \mathcal{D}_{a} & \text{if } \mathcal{D}_{b} < \Delta,\\
  \mathcal{D}_{a} + \mathcal{D}_{b}-\Delta  & \text{if } \mathcal{D}_{b} \geqslant \Delta
  \end{cases}\\
  \mathcal{L}_{seg} =\mathcal{L}_{d} + e^{-1 * \mathcal{L}_{d} * \gamma} * \mathcal{L}_{g},
\end{gather}
where $\gamma$ means a balance parameter to adjust the ratio of $L_{g}$ and $L_{d}$, and $\Delta$ represents the threshold of allowable 
false-positive classification results in exchange for detecting more text features in the low-contrast and indistinguishable area.
\subsection{Attentive Proposal Refinement Module}
Most preliminary detection boxes cover the text instances incompletely, especially those with oriented rectangle shape in the industrial scene. 
To achieve better location accuracy, we propose a novel box selection algorithm by applying attention maps to mine 
high-quality candidate boxes attached to the foreground. 
More foreground boxes are extracted in a high-priority order and fed into the refinement network. 
\subsubsection{Box Selection Algorithm}
Given a set of the prediction boxes with scores $\mathcal{D} =\{(\mathcal{B}_i,\mathcal{S}_i)|i=1,...,l\}$, our goal is to select top $\beta $ boxes and 
apply them into the refinement network. 
First, we binarize the supervised attention map $A$ to obtain the mask map $F$. 
Then, the $F$ will be scaled into the map $F_i$ at each resolution $s_i$ to filter out the invalid anchor points 
and only keep those anchors falling on the predicted foreground regions. 
Specifically, the points with the pixel value of 1 in $F_i$ can be gathered and form the set of $\mathcal{V}  = \{\mathcal{V}_i|i=1,...,l\}$.
Each point in $\mathcal{V} $ corresponds to 8 candidate boxes with different aspect ratios, and the optimal box is selected according to the confidence score.
Therefore, we effectively filter background boxes and obtain a multi-scale candidate box set $\mathcal{\bar{V}} $. 
Finally, the foreground boxes with the top $\beta $ confidence scores are selected from $\mathcal{\bar{V}}$. 
\subsubsection{Refinement Network}
Inspired by \cite{RRPN:2018Arbitrary}, the selected boxes are utilized to extract regions of interest (ROIs) as feature patches from the first four levels in the ResNet-FPN backbone. 
These feature batches are flattened and fed into a fully connected layer to form high-dimensional feature vectors, 
and then two fully connected layers are implemented to predict the classification and regression outputs for each box, respectively.
\subsection{Re-Scoring Mechanism}
For standard post-processing such as Faster R-CNN\cite{fasterRCNN:2017Faster}, Mask R-CNN\cite{MaskR-CNN:2017Mask}, 
the non-maximum suppression (NMS) process is implemented to retain the prediction boxes with the highest score for different objects. 
The confidence scores $S_{c}$ are predicted by the classification branch for each proposal. 
However, this approach may ignore some prediction boxes without the highest classification scores but more accurate locations. 
We thus add an instance score $S_{I}$ to each prediction box as follows:
\begin{gather}
  S_{I} = \frac{\Sigma_{j=1}^{N} \rho_{j}}{N},
\end{gather}
where $\rho_{j}$ represents the pixel value of the attention map $A$.
Compared to directly using weighted instance score $S_{I}$ and confidence score $S_{c}$, 
we adopt the below numeric formulation to form an overall score $S'$,
which has a higher gradient value under the same classification score.
\begin{gather}
  S' = e^{S_{c}}(1+\mu \frac{e^{S_{I}}}{e^{1-S_{I}}}),
\end{gather}
where $\mu$ is the trade-off coefficient. 
Finally, $S'$ is taken as the new confidence score and fed into the NMS algorithm to get the best prediction boxes. 

\subsection{Loss Function}
The overall loss function of RFN consists of $\mathcal{L}_{seg}$, $\mathcal{L}_{det}$, and $\mathcal{L}_{ref}$.
Firstly, the output attention map of the SFF module is optimized by a segmentation loss $\mathcal{L}_{seg}$ under supervised learning to enrich text feature representations. 
Secondly, we calculate the loss of the classification and regression sub-networks (CR Heads) in the one-stage detector according to the following definition to achieve preliminary detection:
\begin{gather}
  \mathcal{L}_{det} = \frac{1}{M}\sum_{i = 1}^{M}(\tau_{i}\mathcal{L}_{reg}(b_{i},b_{i}')+\mathcal{L}_{cls}(s_{i},s_{i}')),  
\end{gather}
where $M$ represents the number of the default boxes. $b_{i}$ and $b_{i}'$ represent an 8-vectors location of the i-th default box 
and prediction box, respectively. $\tau_{i}$ is a binary value indicating whether the i-th default box matches one of the ground-truth 
boxes by IOU. We adopt the focal loss\cite{RetinaNet:2017Focal} for $\mathcal{L}_{cls}$ between the label $s_{i}$ and 
the confidence $s_{i}'$. 
The regression loss $\mathcal{L}_{reg}$ is calculated by the smooth L1 loss\cite{fasterRCNN:2017Faster}. 
Thirdly, $\mathcal{L}_{ref}$ represents the classification and location regression losses of the sampled ROIs generated by APR modules, which is implemented by Faster R-CNN [18]. 
Finally, the total loss is defined as follows:
\begin{gather}
  \mathcal{L}_{total} = \lambda_{1}\mathcal{L}_{seg}+\lambda_{2}\mathcal{L}_{det}+\lambda_{3}\mathcal{L}_{ref},
\end{gather}
where $\lambda_{1}$, $\lambda_{2}$, and $\lambda_{3}$ represent the balance parameter and are set to 1 by default.
\section{Experiments}
In this section, we first evaluate the performance of our method on the MPSC dataset and compare it with the state-of-the-art methods.
We then test RFN on other public scene text datasets and compare them with state-of-the-art methods to demonstrate its robustness.
Finally, we conduct an ablation study of the SFF, APR, and Re-scoring modules on the MPSC dataset. 
\subsection{Implementation Details}
\subsubsection{Evaluation Metrics}
To fairly compare with other methods, we evaluate the proposed method on the MPSC, MSRA-TD500, USTB-SV1K, ICDAR2013 and ICDAR2017-MLT 
datasets, using the standard evaluation protocol proposed in \cite{ICDAR2015:2015ICDAR,MSRATD500:9411951,ICDAR2013:2013ICDAR,ICDAR2017-MLT:2017ICDAR2017,USTB-SV1K}, 
respectively. All experiments are implemented on a server with an NVidia Tesla V100 (32G) GPU.
\subsubsection{Parameters settings}
Our proposed method is optimized by stochastic gradient descent (SGD) with a momentum of 0.9 and a weight decay of 1$\times$10-4. 
The image size is set to 768$\times$768, and the batch size is set to 12. 
A multi-step learning rate strategy is adopted to update weights, in which the initial learning rate is set to 0.001 and halved every 50 epochs. 
In the experiment, $\Delta$ is set to $0.01*S_{gt}$, $\gamma$ is set to $0.1$, $\mu$ is set to $0.5$, and $\beta$ is set to $1000$.  
\begin{table}[htbp]
  \centering
  \caption{Comparison results of text detection of metal parts. 
  }
  \scalebox{1.0}{
  \begin{tabular}{l|cccc} 
  \toprule 
  Algorithms & Precision (\%) & Recall (\%) & F-measure (\%)\\ 
  \midrule 
  EAST \cite{EAST:zhou2017east}  & 76.33 & 73.04 & 74.65 \\
  Mask R-CNN \cite{MaskR-CNN:2017Mask} & 85.28 & 79.25 & 82.15 \\
  RRPN \cite{RRPN:2018Arbitrary} & 81.98 & 78.91 & 80.42 \\ 
  PSENet \cite{PSENET:wang2019shape} & 85.42 & 78.40 & 81.76 \\
  PAN \cite{PAN:9009483} & 87.07 & 81.60 & 84.24 \\
  BDN \cite{BDN:liu2020exploring} & 86.60 & 77.49 & 81.79 \\
  ContourNet \cite{ContourNet:2020ContourNet} & 87.79 & 81.02 & 84.27 \\
  RRPN++ \cite{RRPN++:ma2020rrpn++} & 86.73 & 83.90 & 85.30\\
  FCENet \cite{FCENET:zhu2021fourier} & 87.13 & 81.63 & 84.29\\
  \midrule 
  RFN (ours) & 89.30 & 83.33 & 86.21 \\
  RFN* (ours)  & $\textbf{89.82}$ & $\textbf{84.45}$ & $\textbf{87.05}$ \\
  \bottomrule   
  \end{tabular}
  }
  \par
  \vspace{0.1cm}
  \leftline{\quad All methods except for RFN* are trained on the MPSC dataset.}
  \par
  \leftline{\quad * means pre-training on the SynthMPSC dataset.}
\label{tb:MPSC}
\end{table}
\subsection{Performance Evaluation on MPSC Dataset}
We implement comparative experiments to verify the effectiveness of the proposed method on the MPSC dataset compared with the state-of-the-art 
text detection methods \cite{EAST:zhou2017east,MaskR-CNN:2017Mask,RRPN:2018Arbitrary,PSENET:wang2019shape,PAN:li2018pyramid,BDN:liu2020exploring,
ContourNet:2020ContourNet,RRPN++:ma2020rrpn++,FCENET:zhu2021fourier}.
These methods design novel feature representations and achieve excellent performance in multi-oriented scene text detection. 
As shown in Table \ref{tb:MPSC}, our method achieves the best performance with an F-measure of 86.21\% on the MPSC dataset 
and outperforms the currently best method\cite{RRPN++:ma2020rrpn++} among the nine methods by 1.51\% in the precision metric. 
Moreover, the recall metric of our method ranks only second to RRPN++, which adds an extra recognition branch to recall low-score 
prediction boxes with high recognition scores. 
Note that RFN can also deploy the recognition branch to improve detection performance. 

\begin{figure*}[t]
  \centering
  \includegraphics[width=6.8in]{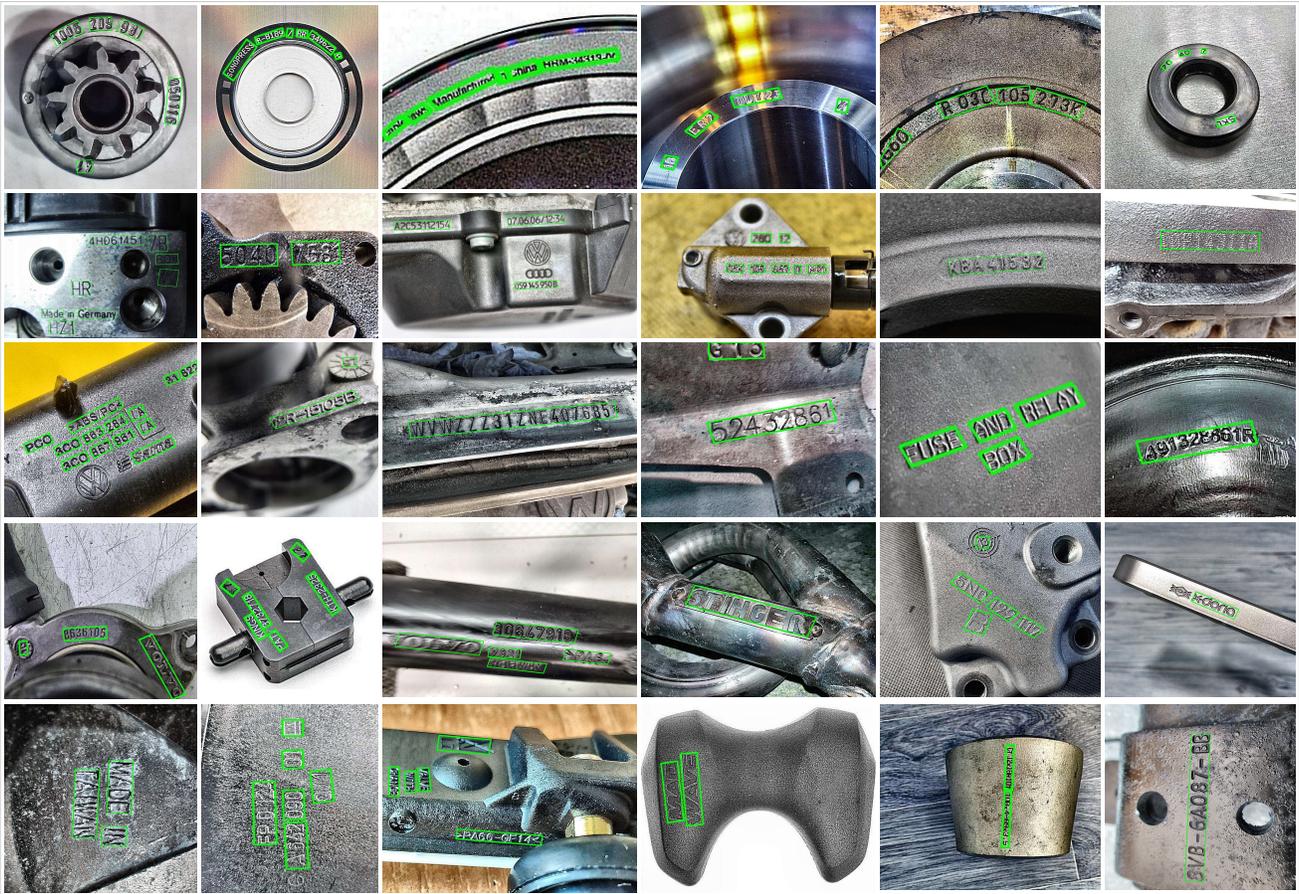}
  \caption{Some multi-oriented detection examples on the MPSC dataset using RFN. Five styles of metal parts text are listed in five rows to illustrate the effectiveness of our method for multi-oriented text detection.}
  \label{Figs.Det_MPSC}
\end{figure*}

To promote the proposed method, we pre-train the RFN network on SynthMPSC dataset and then fine-tune it on the MPSC dataset. 
Specifically, the last row of Table \ref{tb:MPSC} represents the final result, reaching 87.05\% in F-measure. 
Compared to training only on the MPSC dataset, the performance of RFN* improves by 0.84\%, 
verifying that the artificially synthesized samples enhance the ability to detect characters of metal parts. 
Some qualitative results on the MPSC dataset are shown in Fig.\,\,\ref{Figs.Det_MPSC}. 

\begin{figure}[t]
  \centering
  \includegraphics[width=3.5in]{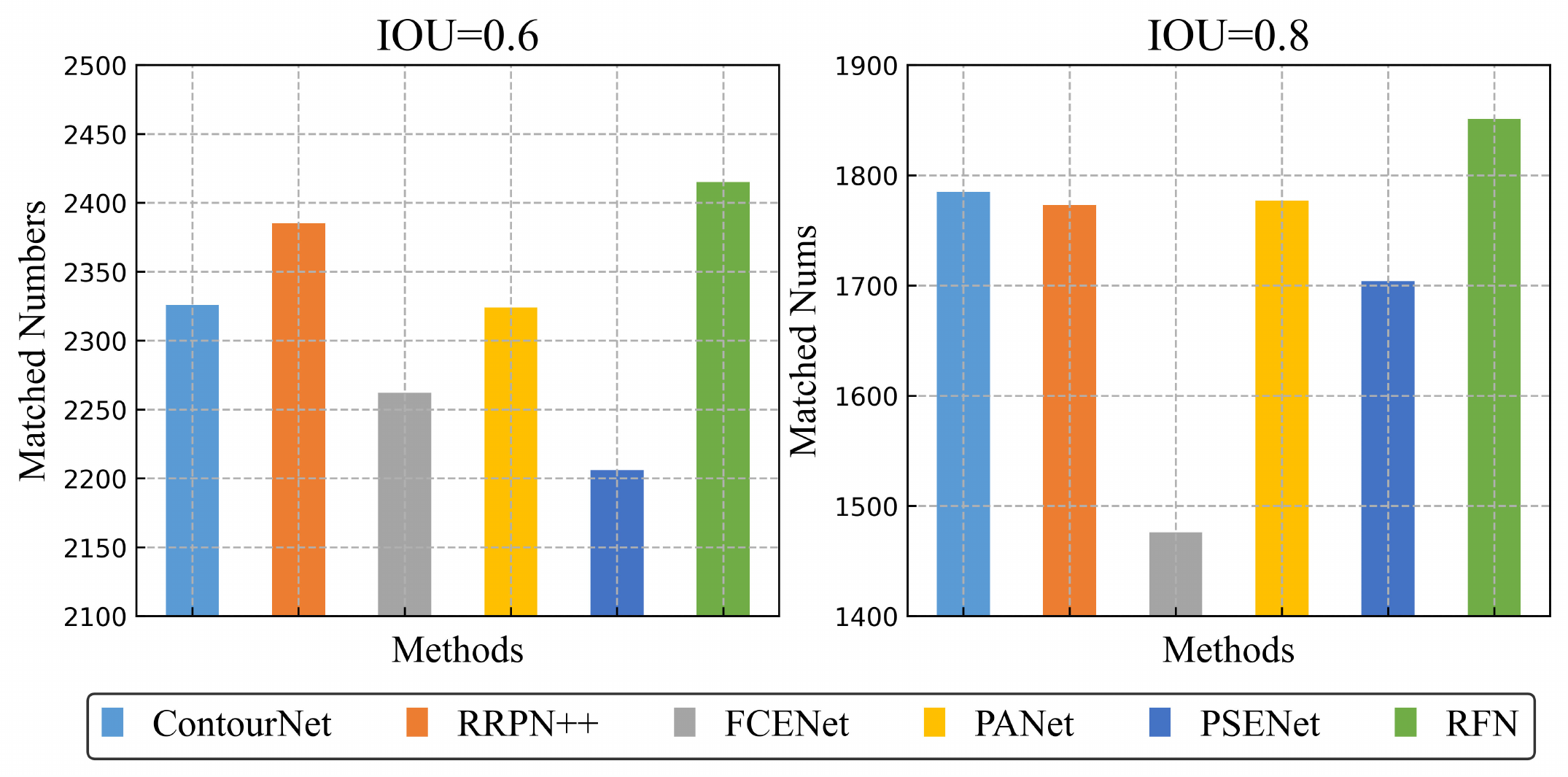}
  \caption{The number of validly matched bounding boxes obtained by different text detection methods in different IOU thresholds. 
  The green bar represents the number of the matched bounding boxes generated by RFN.
  }
  \label{Figs.Match}
\end{figure}

Considering the practical application of deep learning, the quality of text detection results directly determines 
the end-to-end text recognition rate. The accurate bounding boxes provide rich information for the text recognition network. 
Thus different from general evaluation metrics, we change the fixed IOU threshold to calculate the number of matched boxes from the best model of each method. 
Specifically, the predicted bounding box is defined as a matched box if the IOU value between it and 
one of the ground-truth boxes is greater than the artificially set threshold (we set it to 0.6 and 0.8, respectively). 
The number of the matched bounding boxes obtained by different text detection methods is shown in Fig.\,\,\ref{Figs.Match}. 
As represented by green bars, RFN generates the matched bounding boxes with the largest number. 
It implies that the APR module significantly corrects deviations in text localization and generates more high-quality bounding boxes. 
In general, the statistical results demonstrate the effectiveness of our proposed method from another perspective and 
provide substantial improvements for the subsequent text recognition task, which is beneficial to the tracking of metal parts in 
industrial scenes.
\begin{table}[t]
  \centering
  \caption{Comparison results on the MSRA-TD500 dataset. 
  }
  \scalebox{1.0}{
  \begin{tabular}{l|ccc}
   \toprule
   Algorithms &Precision (\%)& Recall (\%)& F-measure (\%)\\
   \midrule
   SegLink$\dagger$ \cite{Seglink:shi2017detecting}  & 86.0 & 70.0 & 77.0\\  
   EAST$\dagger$ \cite{EAST:zhou2017east}  & 87.3 & 67.4 & 76.1\\
   TextSnake$\dagger$ \cite{TextSnake:2018TextSnake}  & 83.2 & 73.9 & 78.3\\
   PixelLink* \cite{Pixellink:deng2018pixellink}  & 83.0 & 73.2 & 77.8 \\
   RRPN \cite{RRPN:2018Arbitrary}  & 82.0 & 68.0 & 74.0 \\
   RRD$\dagger$ \cite{RRD:liao2018rotationsensitive}  & 87.0 & 73.0 & 79.0\\
   Lyu et al. \cite{MOST:lyu2018multioriented}  & 87.6 & 76.2 & 81.5\\
   AS-RPN \cite{ICDAR2013:AS_RPN}  & 84.7 & 80.4 & 82.5 \\
   CRAFT \cite{CRAFT:baek2019character} & 88.2 & 78.2 & 82.9 \\
   ATRR \cite{ATRR:2019Arbitrary}  & 85.2 & 82.1 & 83.6 \\
   PAN$\ddagger$ \cite{PAN:9009483} & 84.4 & 83.8 & 84.1 \\
   \midrule
   RFN (ours) & 88.4  & 80.0 & 84.0\\
   RFN$\ddagger$ (ours) & \textbf{88.4} & \textbf{87.8} & \textbf{88.1}\\
   \bottomrule
  \end{tabular}
  }
  \par
  \vspace{0.1cm}
  \leftline{\quad\, * means training with multiple scales.}
  \par
  \leftline{\quad\, $\dagger$ indicates that the method adds HUST-TR400 dataset\cite{HUST-TD400:yao2014unified} for training.}
  \par
  \leftline{\quad\, $\ddagger$ the blurred text regions labeled as difficult samples are ignored.}
  \label{tb:006}
\end{table}

\begin{figure}[t]
  \centering
  \includegraphics[width=3.5in]{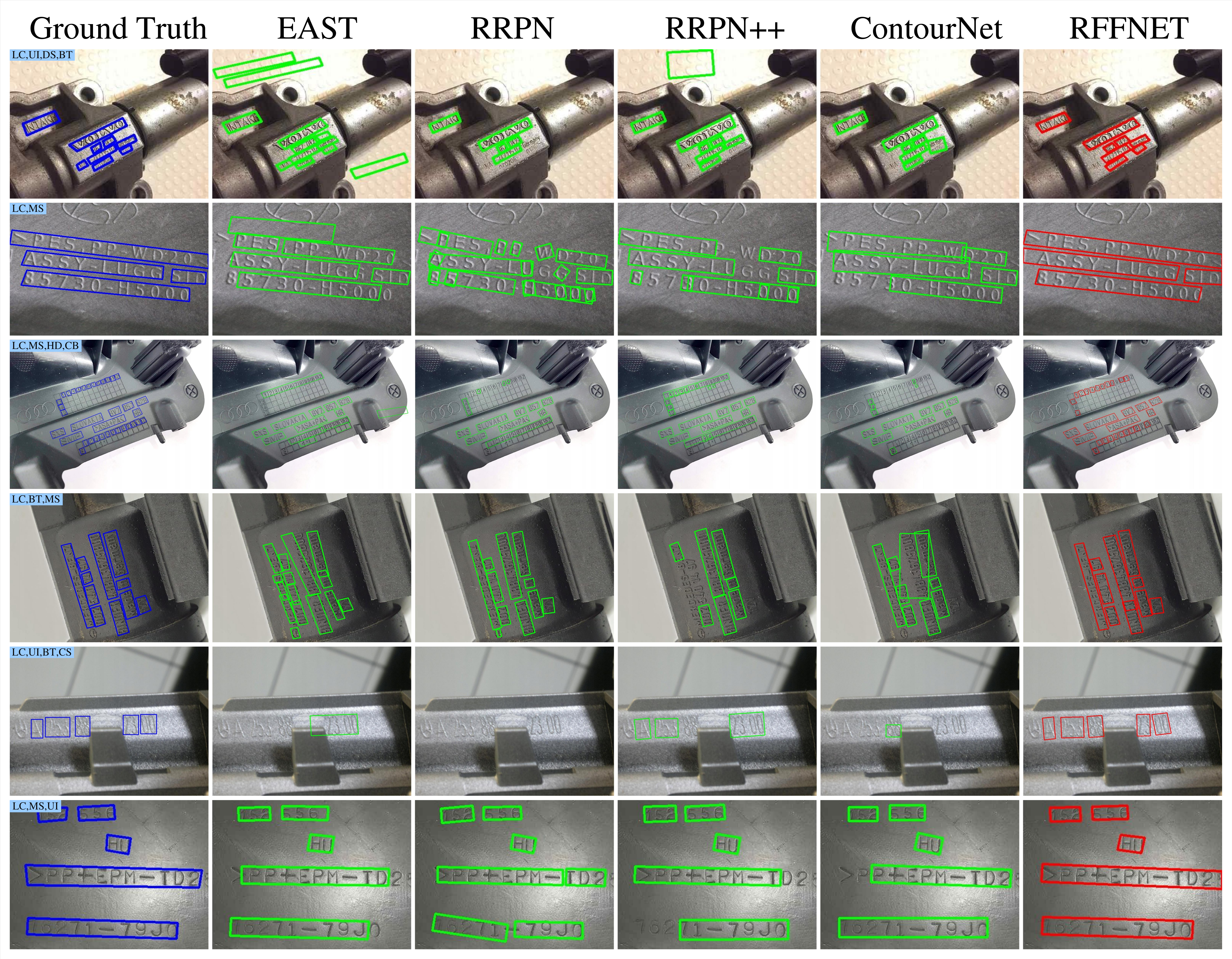}
  \caption{Some comparative examples with RFN and EAST, RRPN, RRPN++, ContourNet methods on the MPSC test set.
  The blue region in the upper left of each image depicts the corresponding challenges.
  }
  \label{Figs.Com}
\end{figure}

Affected by low visual contrast, corroded surfaces, complex backgrounds, etc., 
the texts are not salient and have low visibility in the industrial scene image,  
further limiting the ability of text detection algorithms to be deployed in real-world scenarios. 
Thus we focus on the text foreground feature to weaken the influence of other factors. 
Specifically, SFF guides the framework to obtain more foreground information of metal parts by learning adaptive feature representations. 
The detailed discussions about SFF are given in the following: 
1) Scale-sensitive feature fusion. 
The multi-resolution features are divided into low- and high- levels to enhance the text perception. 
The low-level aggregates foreground features along the channel axis, 
and the exponential operation is adopted to activate the foreground feature response. 
The high-level adaptively learns the features of adjacent subnets by exchanging information mutually, 
which enriches the spatial features and retains more multi-scale text features.
2) Foreground feature optimization. 
We establish a loss mechanism that imposes large weights to focus on foreground prediction results. 
And a threshold of allowable false-positive classification results is set in exchange for 
detecting more text features in the low-contrast and indistinguishable area.
Moreover, the foreground prediction result generated by the SFF module is applied to the entire subsequent localization process, 
including the one-stage detector and refinement network. 
For the detailed discussions, APR is developed to construct high-quality foreground boxes. 
These boxes attached to the foreground are extracted in a high-priority order, 
and many boxes belonging to the background have a low opportunity to be re-corrected, 
which promotes the number of region-of-interest features for bounding box correction.
Some examples of comparison results with other methods are shown in the Fig.\,\,\ref{Figs.Com}.
\subsection{Performance Comparison on Public Datasets}
Our method is evaluated on these typical benchmark datasets to demonstrate its robustness. 
Similar to most scene text detection methods, we also pre-train our method on the SynthText dataset. 
Although SFF and APR modules are designed for text detection of the metal parts, the below experimental results still illustrate 
it gets comparable performance for other scene text detection. 
We report the best results of the comparison methods, each of which was reported in the original paper.
\subsubsection{Detecting Multi-oriented Text}
MSRA-TD500 has become one of the most challenging multi-oriented text datasets with very few training samples and 
super-long large text instances. Covering the text area more completely and appropriately is the biggest challenge.  
Hence we evaluate our method on the MSRA-TD500 benchmark dataset to verify its ability to detect multi-oriented texts.
Table \ref{tb:006} reports our results and compares them with the state-of-the-art 
methods. Our method obtains the best precision and F-measure among all comparison methods. 
Specifically, RFN achieves a precision of 88.4\%, recall of 80.0\%, and F-measure of 84.0\% without extra data training. 
Compared to PAN with ignoring difficult samples to evaluate, our method further achieves 88.1\% in F-measure.
\begin{table}[t]
  \centering
  \caption{Comparison results on the ICDAR2013 dataset. 
  }
  \scalebox{1.0}{
  \begin{tabular}{l|ccc}
   \toprule
   Algorithms & Precision (\%)& Recall (\%)& F-measure (\%)\\
   \midrule
   SegLink* \cite{Seglink:shi2017detecting} & 92.0 & 84.4 & 88.1\\
   SSTD \cite{SSTD:he2017single} & 89.0 & 86.0 & 88.0\\
   TextBoxes++* \cite{Textboxes++:Liao2018TextBoxes}  & 92.0 & 86.0 & 89.0\\
   FOTS \cite{ICDAR2013:liu2018fots} & - & - & 87.3 \\
   RRD* \cite{RRD:liao2018rotationsensitive}  & 92.0 & 86.0 & 89.0\\
   PixelLink* \cite{Pixellink:deng2018pixellink}  & 88.6 & 87.5 & 88.1 \\
   RRPN \cite{RRPN:2018Arbitrary}  & 84.0 & 77.0 & 80.0 \\
   Melinda et al. \cite{ICDAR2013:8977981}& $\textbf{93.9}$ & 91.5 & $\textbf{92.6}$ \\
   FTPN \cite{ICDAR2013:FTPN}  & 93.2 & $\textbf{91.9}$ & 92.5 \\
   Liu et al. \cite{ICDAR2013:liu2019scene}  & 90.2 & 86.3 & 88.2 \\
   Wei et al. \cite{ICDAR2013:9180267}  & 93.7 & 87.4 & 90.4 \\
   \midrule
   RFN (ours)&92.5  &90.7  & 91.6\\
   \bottomrule
  \end{tabular}
  }
  \par
  \vspace{0.1cm}
  \leftline{\quad * means training with multiple scales.}
  \label{tb:008}
\end{table}
\subsubsection{Detecting Horizontal Text}
To verify the robustness to the horizontal texts, we select the popular ICDAR2013 dataset to evaluate the performance of RFN.
Table \ref{tb:008} reports our results and compares them with the state-of-the-art methods. 
It is observed that our result reaches the precision of 92.5\%, recall of 90.7\%, and F-measure of 91.6\%, 
outperforming most existing methods even if some of them are tested in multi-scales. 
It is worth noting that RFN only adopts single-scale images for training and testing, and no other tricks are employed to improve performance. 
It proves that RFN is robust for text detection with horizontal texts in natural scenes.
\subsubsection{Detecting Multi-language Text}
The ICDAR2017-MLT dataset has a wide range of resolutions and includes many small and dense texts from 9 languages.
To fully mine small samples, we use the high resolution to test ICDAR2017-MLT and set the number of candidate boxes $\beta $ to 2500. 
In addition, the ratio aspects of the default boxes are set to \{1, 2, 3, 5, 7.5, 1/2, 1/4, 1/6\} to adapt to the diversity of 
text instances. 
Table \ref{tb:009} reports our results and compares them with the state-of-the-art methods. 
The proposed method achieves competitive performance with an F-measure of 73.0\%. 
Although each language has its specificity, RFN can successfully locate the text instances with multiple languages, 
demonstrating the robustness in multilingual scenes.
\begin{table}[t]
  \centering
  \caption{Comparison results on the ICDAR2017-MLT dataset. 
  }
  \scalebox{1.0}{
  \begin{tabular}{l|ccc}
   \toprule
   Algorithms &Precision (\%)& Recall (\%)& F-measure (\%)\\
   \midrule
   Sensetime OCR \cite{ICDAR2017-MLT:2017ICDAR2017} & 56.9 & 69.4 & 62.6\\
   FOTS \cite{ICDAR2013:liu2018fots} & 81.0 & 57.5 & 67.3 \\
   FOTS* \cite{ICDAR2013:liu2018fots} & 81.9 & 62.3 & 67.3 \\
   LOMO \cite{LOMO:zhang2019look} & 78.8 & 60.6 & 68.5\\
   LOMO* \cite{LOMO:zhang2019look} & 80.2 & 67.2 & 73.1\\
   PSENet \cite{PSENET:wang2019shape} & 73.8 & 68.2 & 70.9 \\
   PSENet$\ddagger$ \cite{PSENET:wang2019shape} & 75.4 & 69.2 & 72.1 \\
   CharNet \cite{ICDAR2017MLT:xing2019convolutional} & 77.1 & $\textbf{70.1}$ & 73.4 \\
   CRAFT \cite{CRAFT:baek2019character} & 80.6 & 68.2 & 73.9 \\
   Unrealtext \cite{Unrealtext:long2020unrealtext} & $\textbf{82.2}$ & 67.4 &$\textbf{74.1}$ \\
   ISNet \cite{ISNet} & 78.0 & 67.4 & 72.3\\
   \midrule
   RFN (ours) & 79.4 & 67.6 & 73.0\\
   \bottomrule
  \end{tabular}
  }
  \par
  \vspace{0.1cm}
  \leftline{\; * means training with multiple scales.}
  \par 
  \leftline{\; $\ddagger$ means the model uses ResNet152 as the backbone.}
  \label{tb:009}
\end{table}
\subsubsection{Detecting Low-resolution Text}
To further evaluate the generalization ability of RFN, we select the challenging USTB-SV1K dataset with many low 
resolution and blurred images. Table \ref{tb:007} reports our results and compares them with the state-of-the-art methods. 
Without pre-training on the SynthText dataset, our method still achieves the best results, 
outperforming Wang et al.'s method (the best-reported result currently) by 3.2\% in the F-measure metric. 

Overall, expensive experiments illustrated that our method robustly detects multi-oriented texts, horizontal texts, 
multi-language texts, low-resolution texts, and can be deployed in more complex scenes.
\begin{table}[t]
  \centering
  \caption{Comparison results on the USTB-SV1K dataset.}
  \scalebox{1.0}{
  \begin{tabular}{l|ccc}
   \toprule
   Algorithms & Precision (\%)& Recall (\%)& F-measure (\%)\\
   \midrule
   Liu et al.$\dagger$ \cite{ICDAR2013:liu2019scene} & 72.3 & 50.3 & 59.3\\
   FTPN \cite{ICDAR2013:FTPN}  & 61.4 & 63.8 & 62.6 \\
   Wang et al.$\dagger$ \cite{USTB-SV1K:Wang}  & 73.0 & 67.0 & 70.0\\
   \midrule
   RFN (ours) & $\textbf{80.3}$ & $\textbf{67.2}$ & $\textbf{73.2}$\\
   \bottomrule
  \end{tabular}
  }
  \par
  \vspace{0.1cm}
  \leftline{\quad\, $\dagger$ indicates that the method adds extra data to train.}
  \label{tb:007}
\end{table}
\subsection{Ablation Study}
We implement an ablation experiment for our method on the MPSC dataset. 
Specifically, we analyze the influence of the model structure on performance for the MPSC dataset.
The model is split into five combinations and trained separately to verify the effectiveness of the proposed method. 
The comparative test results are shown in Table \ref{tb:Model Structure}.
\subsubsection{Segmentation-based Foreground-focus Module (SFF)}
The SFF module improves the accuracy rate and recall rate by 2.78\% and 3.43\%, respectively. 
It proves that the foreground-focus mask branch is suitable for the MPSC dataset and effectively highlights the text features 
in the complex background of metal parts. 
Compared to the baseline, the mask branch integrates the high-quality attention map into the regression network to generate discriminative feature representations, which optimizes the geolocation parameters and provides more competitive prediction boxes. 
The attention map is the primary factor in improving text detection performance in the MPSC dataset.
\begin{figure}[t]
  \centering
  \includegraphics[width=3.5in]{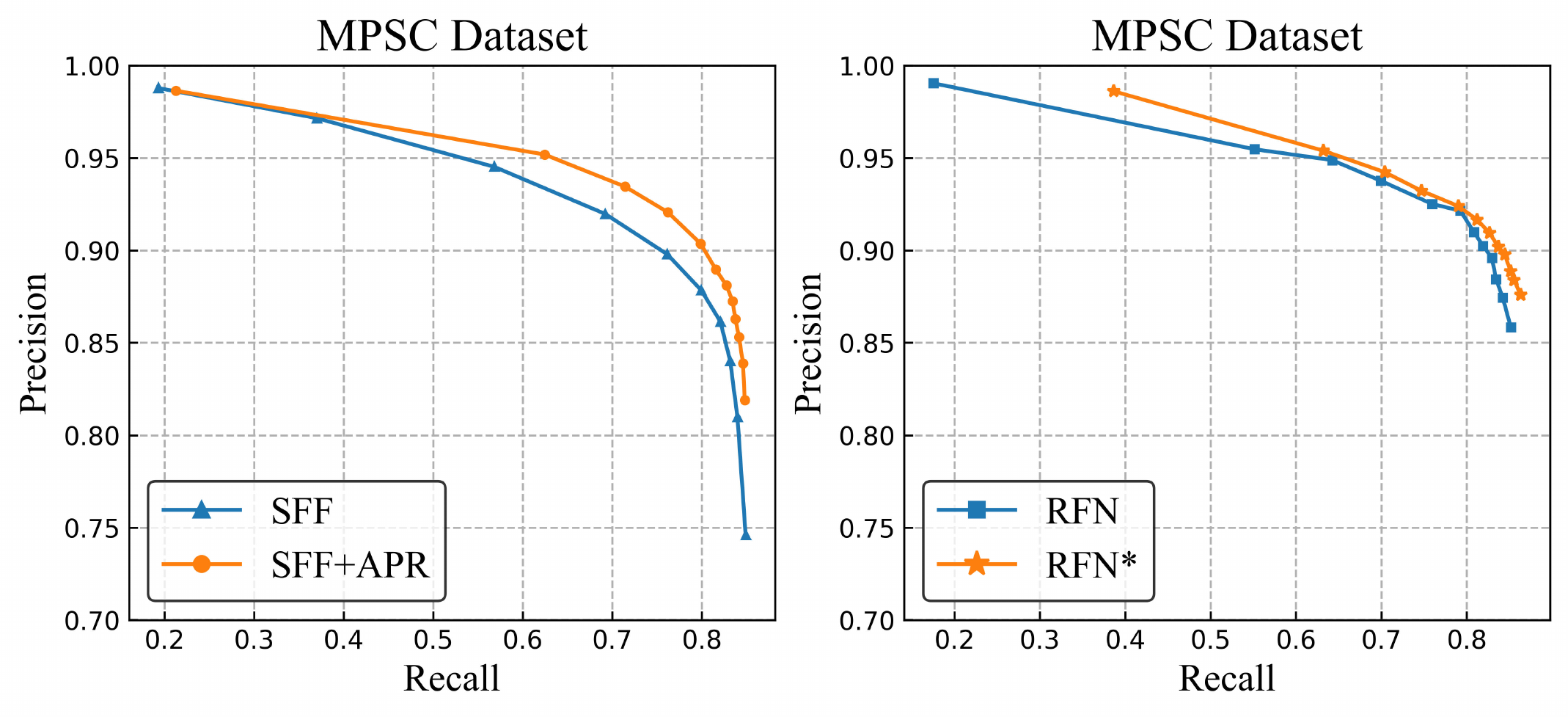}
  \caption{Precision-recall curves. 
  We evaluate the robustness of the two combinations (``SFF'' \emph{vs} ``SFF+APR'', ``RFN'' \emph{vs} ``RFN*''), separately.
  }
  \label{Figs.PR}
\end{figure}
\begin{table}[t]
  \centering
  \caption{Evaluate the effectiveness of the MPSC dataset in the proposed modules of SFF, APR, Re-score.}
  \scalebox{0.8}{
  \begin{tabular}{ccc|ccc|c}
    \toprule 
      SFF & APR & Re-score & Precision (\%)& Recall (\%)& F-measure (\%)& $\Delta$ F (\%)\\
      \midrule  & & & 82.41 & 79.22 & 80.78 & $-$ \\
        \Checkmark &            &            & 85.19 & 82.65 & 83.90 & 3.12\% \\
        \Checkmark &            & \Checkmark & 85.44 & 83.09 & 84.25 & 3.47\% \\
        \Checkmark & \Checkmark &            & 89.18 & 83.19 & 86.08 & 5.30\% \\
        \Checkmark & \Checkmark & \Checkmark & 89.30 & 83.33 & 86.21 & 5.43\% \\
        \bottomrule      
      \end{tabular}
      }
      \par
      \vspace{0.1cm}
      \leftline{\quad $\Delta$F is the improvement of F-measure relative to baseline.}
      \label{tb:Model Structure}
\end{table}
\subsubsection{Attentive Proposal Refinement Module (APR)}
The combination of SFF and APR further improves the text detection performance, reaching 86.08\% in F-measure. 
It shows that APR significantly improves the precision metric of detection performance, which first applies the attention map 
to mine candidate boxes in complex backgrounds of metal parts. 
On the one hand, these boxes attached to the foreground are extracted in a high-priority order and fed into the refinement network. 
It means that many boxes belonging to the background have a low opportunity to be re-corrected, reducing the number of false positives (FP). 
On the other hand, multi-scale detection boxes with high classification scores in the first stage are selected to improve the quality of 
candidate boxes. Ideally, a good candidate box can accurately cover the text area, so there is no need to re-correct the location deviation. 
Therefore, high-quality candidate boxes promote the refinement network to generate more accurate prediction boxes with high IOU values 
to ground-truth boxes. 
It increases the number of true positives (TP). 
As shown in Fig.\,\,\ref{Figs.Match}, RFN generates more prediction bounding boxes with high IOU. 
The increase in TP and the decrease in FP improve the precision metric. 
\subsubsection{Re-scoring Mechanism}
The re-scoring module responds to more accurate prediction boxes by adding instance scores and 
positively affects all combinations in Table \ref{tb:Model Structure}.
Some prediction boxes with low classification scores and high instance scores are kept as final bounding boxes 
to increases the number of true positives (TP).
Compared to SFF+APR, the re-scoring module has a more significant impact on SFF, side reflecting that APR improves the overall 
location accuracy of the prediction boxes. 
More importantly, the precise location can improve text recognition performance and facilitate metal parts tracking while maintaining measurement indicators. 

Finally, we draw the precision-recall curve of text detection on the MPSC dataset as shown in Fig. \ref{Figs.PR} to illustrate 
the whole performance of the model.
\subsubsection{Recognition Head}
To prove the effectiveness of the RFN method, we add two groups of control experiments. 
First, we cancel the recognition head of RRPN++. As shown in the second and fourth rows of Table \ref{tb:Recognition Structure}, the recall rate is 1.36\%
lower than that of RFN, which demonstrates that our method achieves the best performance among the comparative methods above. 
Then the recognition branch is added to our proposed method to improve the detection performance. 
As shown in the third and fifth rows of Table \ref{tb:Recognition Structure}, 
our recall rate is 0.52\% higher than RRPN++ and exceeds its accuracy metric by 2.73\%.
\begin{figure}[t]
  \centering
  \includegraphics[width=3.4in]{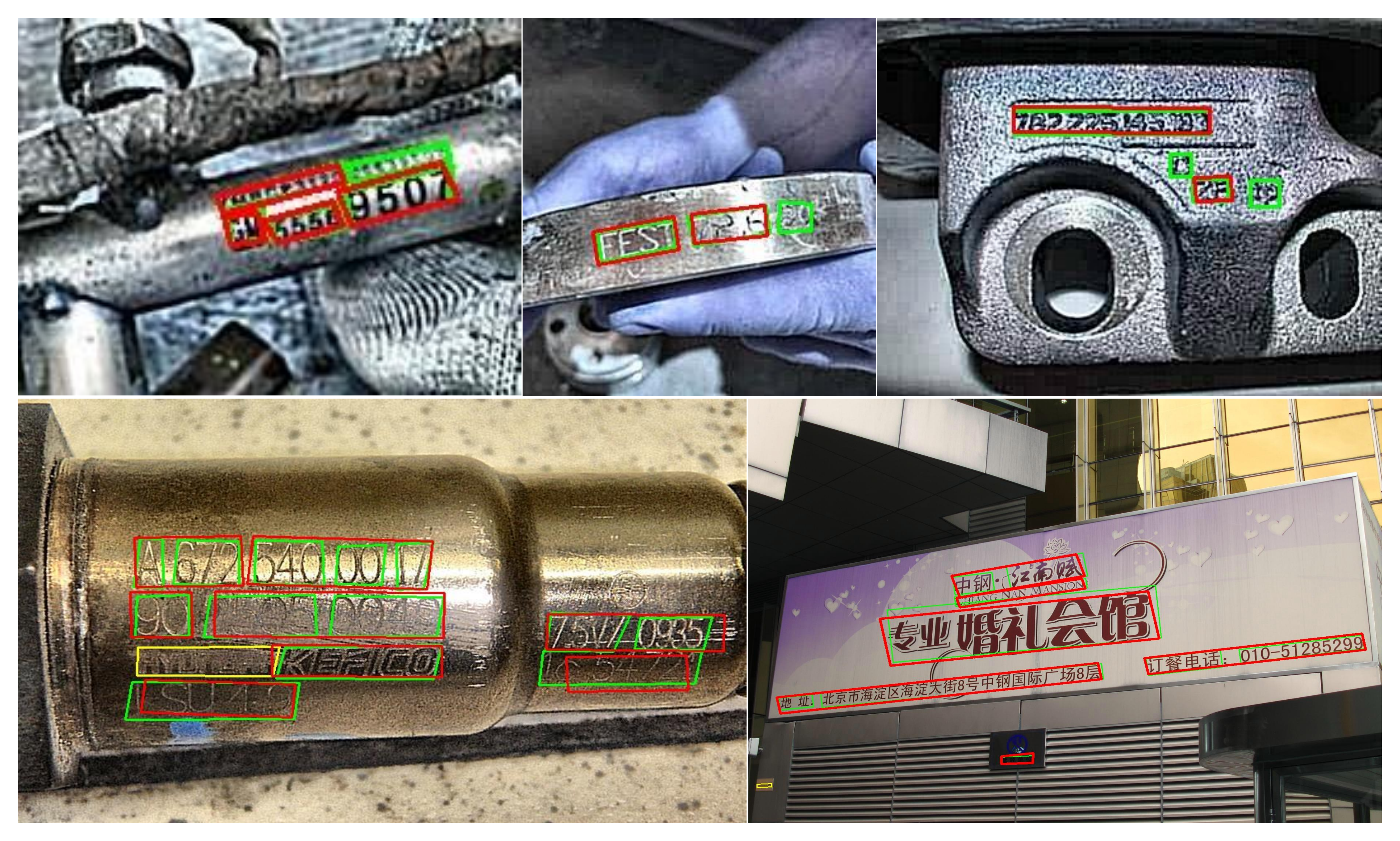}
  \caption{Fail examples of text detection results generated by RFN. The green bounding boxes are the labels, the red bounding boxes are generated by our proposed method, and the yellow boxes are ignored in training stage (the text label is '\#\#\#').}
  \label{Figs.13}
\end{figure}

\begin{table}[t]
  \centering
  \caption{Ablation on efficiency of Recognition branch.}
  \scalebox{0.9}{
  \begin{tabular}{c|c|ccc}
    \toprule 
      Method & Rec. & Precision (\%)& Recall (\%)& F-measure (\%)\\
    \midrule  
      \multirow{2}{*}{RRPN++} & \XSolidBrush & 86.21 & 81.97 & 84.03 \\
                              & \Checkmark & 86.73 & 83.90 & 85.30 \\
    \midrule 
      \multirow{2}{*}{RFN} & \XSolidBrush & 89.30 & 83.33 & 86.21 \\
                           & \Checkmark & 89.46  & 84.42  & 86.87  \\
    \bottomrule      
  \end{tabular}
}
  \label{tb:Recognition Structure}
\end{table}
\subsection{Limitation}
Some failure examples cause performance reduction.
Specifically, one class of failed examples happened at extremely low-resolution industrial images 
where our RFN method exists false negatives in the low-salient text regions as shown in the first row of Fig.\,\,\ref{Figs.13}. 
Another class listed in the second row is mislocated sentence-level and word-level texts affected by the spacing between words 
due to a non-unified standard for the spacing between sentences, words, and characters among the various labels.
\section{Conclusion}
In this paper, 
we propose a effective text detection method, RFN, to locate text instances of metal parts in industrial scenes. 
By designing the SFF, APR, and re-scoring modules, RFN is robust to tackle location deviation problems in the complex background. 
Experiments demonstrate that our method achieves the state-of-the-art performance on the MPSC dataset.
Second, our method effectively detects multi-oriented, horizontal, and multi-language texts and gets competitive performance 
on public benchmark datasets, indicating the generalization ability to be deployed in more complex scenes.
Third, we contribute two benchmark datasets of metal parts (MPSC and SynthMPSC dataset) dedicated to industrial text detection research. 
To the best of our knowledge, these are the first industrial text datasets. 
In the future, we will apply text detection in metal parts tracking, involving a text recognition task to record metal parts 
information in industrial production lines. 
% if have a single appendix:
%\appendix[Proof of the Zonklar Equations]
% or
%\appendix  % for no appendix heading
% do not use \section anymore after \appendix, only \section*
% is possibly needed

% use appendices with more than one appendix
% then use \section to start each appendix
% you must declare a \section before using any
% \subsection or using \label (\appendices by itself
% starts a section numbered zero.)
%

% \appendices
% \section{Proof of the First Zonklar Equation}
% Appendix one text goes here.

% % you can choose not to have a title for an appendix
% % if you want by leaving the argument blank
% \section{}
% Appendix two text goes here.

% % use section* for acknowledgment
% \section*{Acknowledgment}

% The authors would like to thank...

% Can use something like this to put references on a page
% by themselves when using endfloat and the captionsoff option.
\ifCLASSOPTIONcaptionsoff
  \newpage
\fi

\begin{IEEEbiography}[{\includegraphics[width=1in,height=1.25in,clip,keepaspectratio]{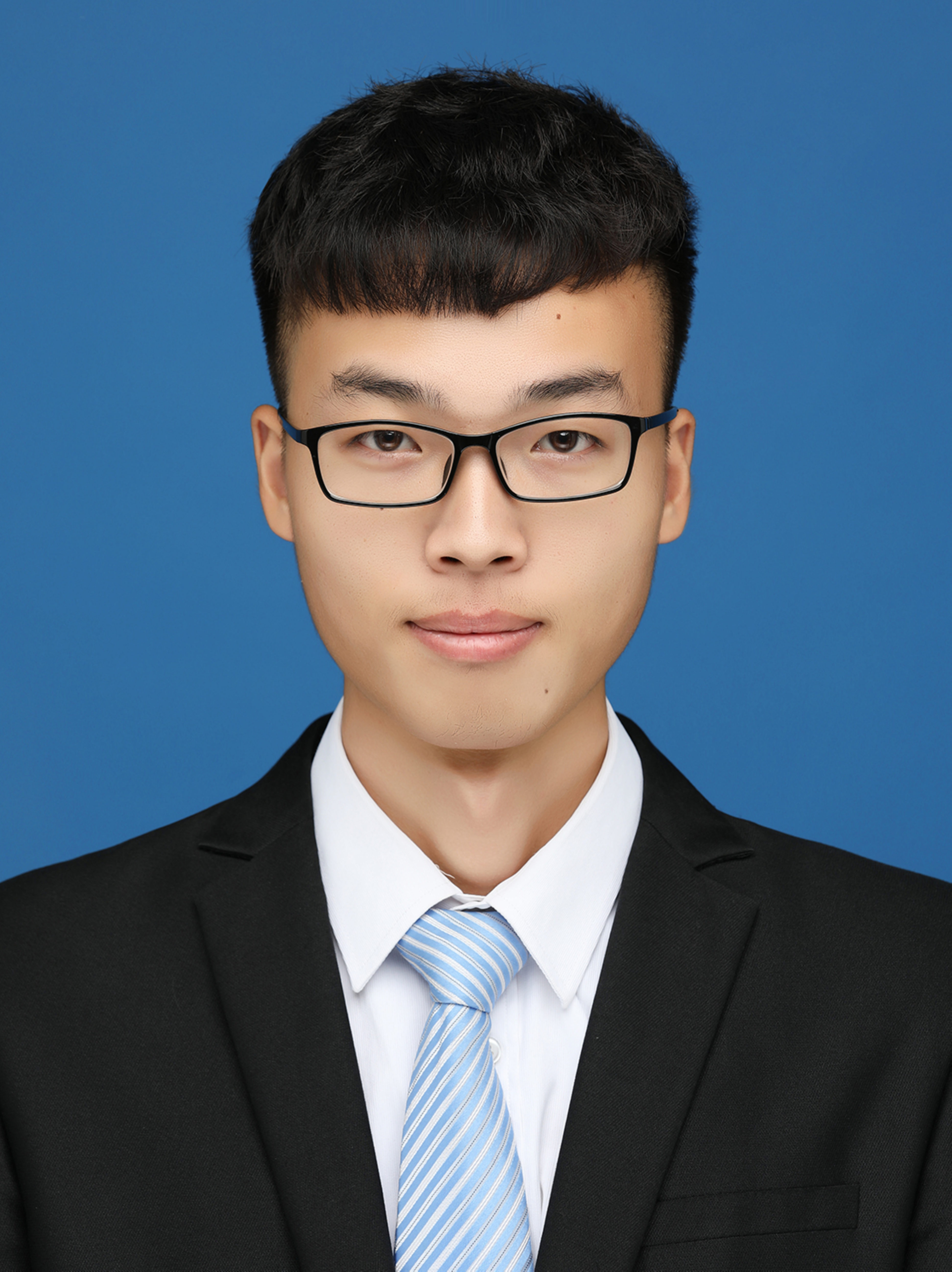}}]{Tongkun Guan}
  % or if you just want to reserve a space for a photo:
  received the B.S. degree in Electrical Engineering and Automation from Hunan University, Changsha, China, in 2020. 
  Currently, he is an M.S. student with the Key Laboratory of System Control and Information Processing, 
  Shanghai Jiao Tong University. He has wide research interests mainly including computer vision, text detection, 
  image processing, and text recognition. 
    \end{IEEEbiography}
% \vfill
% \newpage
\begin{IEEEbiography}[{\includegraphics[width=1in,height=1.25in,clip,keepaspectratio]{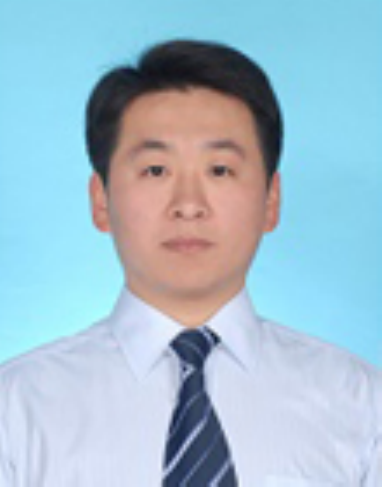}}]{Chaochen Gu}
  % or if you just want to reserve a space for a photo:
  is currently an Associate Professor at School of Electronic Information and Electrical Engineering, Shanghai Jiao Tong University. 
  He received his bachelor degree from Shandong University, Jinan, China, in 2007, 
  and the Ph.D. degree in Mechanical Engineering from Shanghai Jiao Tong University, Shanghai, China, in 2013. 
  His current research interests include industry robotics, machine vision, and man-machine interfaces.
  \end{IEEEbiography}
\begin{IEEEbiography}[{\includegraphics[width=1in,height=1.25in,clip,keepaspectratio]{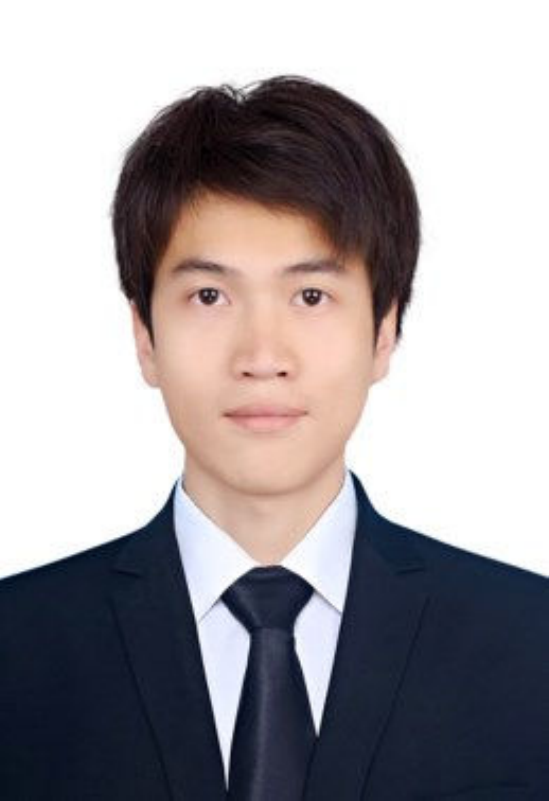}}]{Changsheng Lu}
  % or if you just want to reserve a space for a photo:
  is currently a Ph.D. student with the College of Engineering and Computer Science at The Australian National University. He received the M.S. degree in control science and engineering from Shanghai Jiao Tong University, Shanghai, China, in 2020, and received the B.S. degree in Automation from Southeast University, Nanjing, China, in 2017. He has wide research interests mainly including computer vision, few-shot learning, transfer learning, image processing, pattern recognition, and robotics. Particularly, he is interested in the theories and algorithms that empower robot to see, think and conduct more like a human. Previously, he was awarded the national scholarship and the outstanding graduate student of SEU and SJTU. He has served as the reviewers of IJCV, IEEE T-IP, IEEE Computational Intelligence Magazine, PR, IEEE RA-L, and JVCIR.
  \end{IEEEbiography}
\begin{IEEEbiography}[{\includegraphics[width=1in,height=1.25in,clip,keepaspectratio]{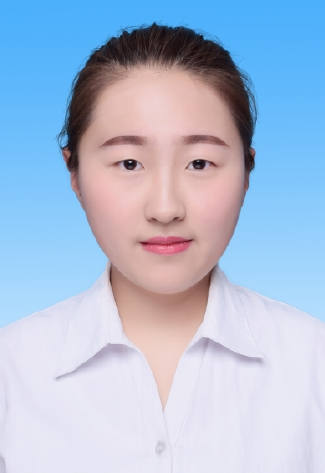}}]{Jingzheng Tu}
  % or if you just want to reserve a space for a photo:
  received the B.Eng. degree in the college of automation from Xi'an Jiaotong University, Xi'an, China, in 2018. 
  She is currently pursuing the Ph.D. degree at the Department of Electronic Engineering, Shanghai Jiao Tong University, Shanghai, 
  China. Her current research interests include intelligent video analytics of edge-enabled industrial Internet-of-Things.
  \end{IEEEbiography}
\begin{IEEEbiography}[{\includegraphics[width=1in,height=1.25in,clip,keepaspectratio]{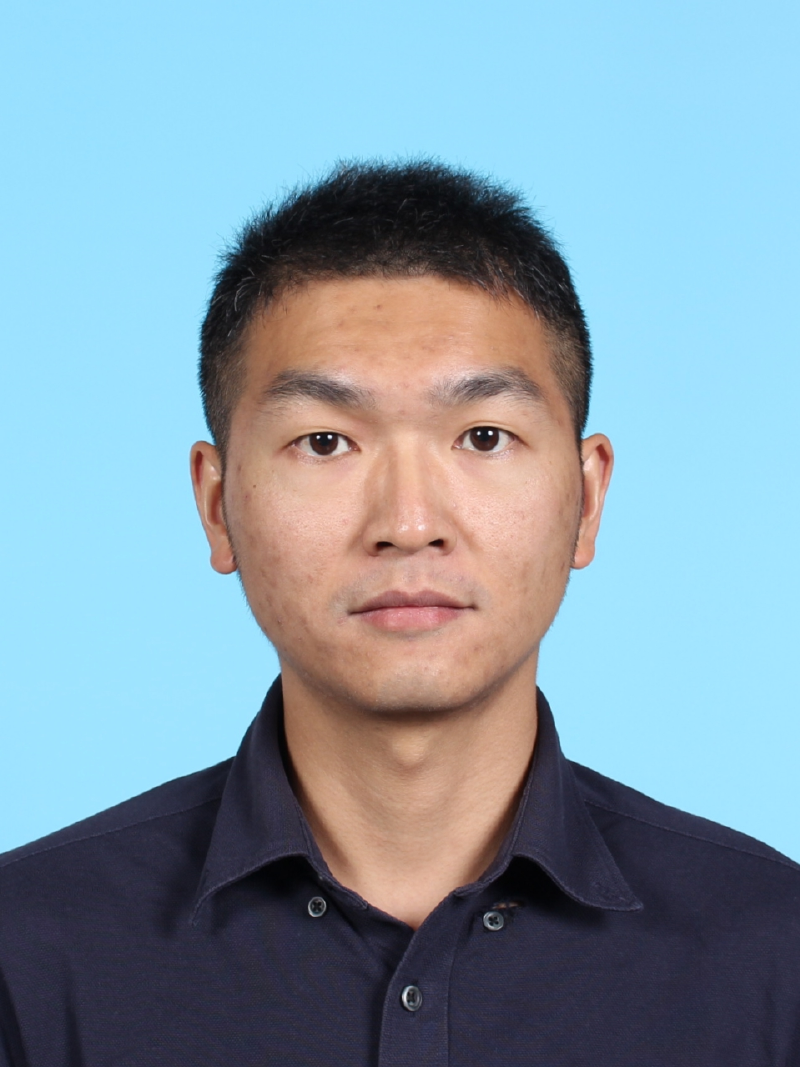}}]{Qi Feng}
  % or if you just want to reserve a space for a photo:
  received the B.S. degree in automation from the Nanjing University, China, in 2010. 
  He is currently working toward the Ph.D. degree in the School of Electronic Information and Electrical Engineering, 
  the Shanghai Jiao Tong University. His research interests include computer vision, machine learning, and 3-D scene understanding.
  \end{IEEEbiography}
\begin{IEEEbiography}[{\includegraphics[width=1in,height=1.25in,clip,keepaspectratio]{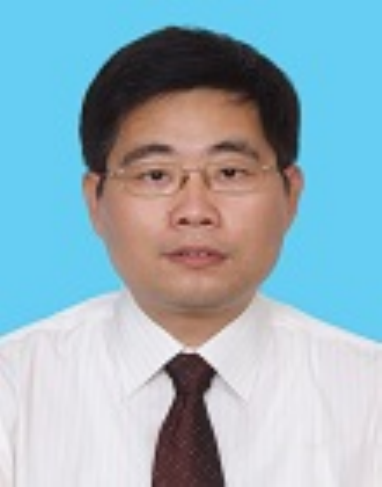}}]{Kaijie Wu}
  % or if you just want to reserve a space for a photo:
  is an Associate Professor at School of Electronic Information and Electrical Engineering, Shanghai Jiao Tong University. 
  He received his Ph.D. degree in Biomedical Engineering from Tianjin University, Tianjin, China, in 2006. 
  His current research explores biomedical optical imaging, medical information processing, and pattern recognition.
  \end{IEEEbiography}
\begin{IEEEbiography}[{\includegraphics[width=1in,height=1.25in,clip,keepaspectratio]{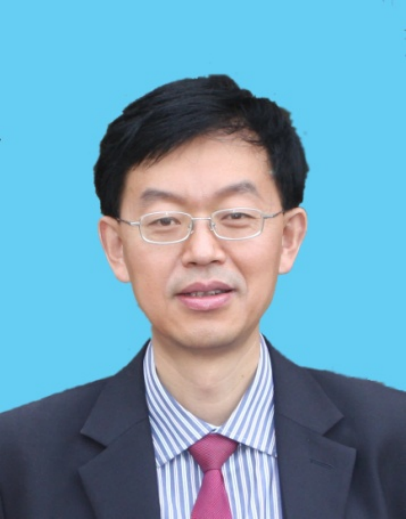}}]{Xinping Guan}
  % or if you just want to reserve a space for a photo:
  Xinping Guan (M'02-SM'04-F'18) received B.Sc. degree in Mathmatics from Harbin Normal University, 
  China in 1986 and PhD degree in Control and Systems from Harbin Institute of Technology, China in 1999. 
  He is currently a Chair Professor with Shanghai Jiao Tong University, Shanghai, China, 
  where he is the Deputy Director of the University Research Management Office 
  and the Director of the Key Laboratory of Systems Control and Information Processing, Ministry of Education of China. 
  His current research interests include industrial cyber-physical systems and wireless networking and applications 
  in smart factory. He is the Leader of the prestigious Innovative Research Team, 
  National Natural Science Foundation of China (NSFC).
  Dr. Guan is an Executive Committee Member of the Chinese Automation Association Council 
  and the Chinese Artificial Intelligence Association Council. 
  He was the recipient of the First Prize of Natural Science Award from the Ministry of Education of China in both 2006 and 2016, 
  and the Second Prize of the National Natural Science Award of China in 2008. 
  He is a “National Outstanding Youth” honored by the NSFC, and the “Changjiang Scholar” by the Ministry of Education of China.
  \end{IEEEbiography}
\vfill
% that's all folks
\end{document}